%% file: _main.tex
\input{_constants}
\arxiv
\pdfoutput=1
\documentclass[10pt,twocolumn,letterpaper]{article}
\input{cvpr_header}

\unless\ifarxiv \myexternaldocument{_supplementary} \fi

\usepackage{tabu}                     
\usepackage{multirow}                 
\usepackage{multicol}                 
\usepackage{multirow}                
\usepackage{float}                    
\usepackage{makecell}                 
\usepackage{booktabs}                 

\usepackage{indentfirst}
\begin{document}
\title{\paperTitle}
\author{\authorBlock}
\maketitle

\input{00_abstract}
\input{01_intro}
\input{02_related}

\input{03_method}

\input{04_experiments}

\input{10_conclusion}
\clearpage
{\small
\input{main.bbl}

}
\ifarxiv \clearpage \appendix \input{12_appendix} \fi
\end{document}

%% file: _constants.tex
\def\paperID{14146}
\def\confName{ICCV}
\def\confYear{2025}

\def\paperTitle{
MSNeRV: Neural Video Representation with Multi-Scale Feature Fusion
}

\def\authorBlock{
    Jun Zhu \qquad
    Xinfeng Zhang\qquad
    Lv Tang \qquad
    Junhao Jiang \\
    University of Chinese Academy of Sciences\\
    {\tt\small \{zhujun23, xfzhang\}@mails.ucas.ac.cn,}\\
    {\tt\small luckybird1994@gmail.com,}\\
    {\tt\small 2021212205134@stu.hznu.edu.cn}\\
}

\newif\ifreview 
\newif\ifarxiv \newcommand{\arxiv}{\arxivtrue}
\newif\ifcamera 
\newif\ifrebuttal 

%% file: cvpr_header.tex
\ifreview \usepackage[review]{cvpr} \fi
\ifarxiv \usepackage[pagenumbers]{cvpr} \fi
\ifrebuttal \usepackage[rebuttal]{cvpr} \fi
\ifcamera \usepackage{cvpr} \fi

\input{_macros}  

\usepackage{xr-hyper}

\makeatletter
\newcommand*{\addFileDependency}[1]{
  \typeout{(#1)}
  \@addtofilelist{#1}
  \IfFileExists{#1}{}{\typeout{No file #1.}}
}

\makeatother
\newcommand*{\myexternaldocument}[1]{
    \externaldocument{#1}
    \addFileDependency{#1.tex}
    \addFileDependency{#1.aux}
}

\definecolor{cvprblue}{rgb}{0.21,0.49,0.74}
\usepackage[pagebackref,breaklinks,colorlinks,allcolors=cvprblue]{hyperref}
\usepackage[capitalize]{cleveref}
\crefname{section}{Sec.}{Secs.}
\crefname{table}{Table}{Tables}
\crefname{figure}{Fig.}{Figs.}

\ifarxiv \crefname{appendix}{App.}{Apps.}
\else \crefname{appendix}{Suppl.}{Suppls.} \fi

%% file: _macros.tex
\usepackage{graphicx}	
\usepackage{amsmath}	
\usepackage{amssymb}	
\usepackage{booktabs}
\usepackage{times}
\usepackage{microtype}
\usepackage{epsfig}
\usepackage{caption}
\usepackage{float}
\usepackage{placeins}
\usepackage{color, colortbl}
\usepackage{stfloats}
\usepackage{enumitem}
\usepackage{tabularx}
\usepackage{xstring}
\usepackage{multirow}
\usepackage{xspace}
\usepackage{url}
\usepackage{subcaption}
\usepackage{xcolor}
\usepackage[hang,flushmargin]{footmisc}

\ifcamera \usepackage[accsupp]{axessibility} \fi

% ...

\ifarxiv  \fi

\newcommand{\R}[1]{{%
    \textbf{%
        \ifstrequal{#1}{1}{\textcolor{red}{R#1}}{%
        \ifstrequal{#1}{2}{\textcolor{blue}{R#1}}{%
        \ifstrequal{#1}{3}{\textcolor{magenta}{R#1}}{%
        \ifstrequal{#1}{4}{\textcolor{teal}{R#1}}{%
                           \textcolor{cyan}{R#1}%
        }}}}%
    }%
}}

%% file: 00_abstract.tex
\begin{abstract}
Implicit Neural representations (INRs) have emerged as a promising approach for video compression, and have achieved comparable performance to the state-of-the-art codecs such as H.266/VVC. However, existing INR-based methods struggle to effectively represent detail-intensive and fast-changing video content. This limitation mainly stems from the underutilization of internal network features and the absence of video-specific considerations in network design. To address these challenges, we propose a multi-scale feature fusion framework, MSNeRV, for neural video representation. 
In the encoding stage, we enhance temporal consistency by employing temporal windows, and divide the video into multiple Groups of Pictures (GoPs), where a GoP-level grid is used for background representation.
Additionally, we design a multi-scale spatial decoder with a scale-adaptive loss function to integrate multi-resolution and multi-frequency information.
To further improve feature extraction, we introduce a multi-scale feature block that fully leverages hidden features. 
We evaluate MSNeRV on HEVC ClassB and UVG datasets for video representation and compression. Experimental results demonstrate that our model exhibits superior representation capability among INR-based approaches and surpasses VTM-23.7 (Random Access) in dynamic scenarios in terms of compression efficiency. 
\end{abstract}

%% file: 01_intro.tex
\section{Introduction}
\label{sec:intro}
With the advancement of modern multimedia, video compression has become increasingly important. 
Traditional hybrid video coding \cite{wien2015high,bross2021overview,wiegand2003overview} has evolved into a mature standard over decades. However, its modular design may lead to suboptimal synergy between components. In response, deep learning-based approaches \cite{lu2019dvc,li2021deep,li2023neural,li2024neural} have been integrated into video coding. These methods can extract spatiotemporal features based on leaning from large datasets and achieve global optimality through end-to-end optimization. Some learning-based video compression methods (such as DCVC-DC \cite{li2023neural}) have surpassed the latest conventional video coding standard, H.266/VVC \cite{bross2021overview,moore1991vtm}. However, learning-based methods often suffer from limited generalization ability. For instance, the DCVC series codecs \cite{li2021deep,li2023neural,li2024neural} exhibit significant performance degradation when processing screen content videos \cite{tang2025canerv}, likely due to the absence of such data in training datasets.

\input{epoch}
To address this problem, methods such as NeRV \cite{chen2021nerv} utilize implicit neural representations (INRs) to model video sequences holistically. 
INR-based methods do not rely on a fixed prior learned from training data.
They treat videos as functions that map spatiotemporal coordinates to pixel values. 
A neural network is trained to approximate this mapping, with its parameters serving as a compact representation for video storage and reconstruction.
Their compression performance is primarily influenced by the model's representation capacity and inherent characteristics of the video.
HNeRV \cite{chen2023hnerv} improves coding efficiency by introducing an explicit feature storage structure, while HiNeRV \cite{kwan2024hinerv} further advances the approach with hierarchical encoding.

However, these INR-based methods fail to match the performance of other compression approaches on highly detailed and dynamic videos.  
This is due to the model's limited representational capability, which prevents it from capturing complex spatiotemporal correlations.
A model's representational capability is tied to its size, but simply scaling up the model leads to higher compression bitrates. Thus, the challenge is to enhance the model's learning ability without increasing its size.
The problem becomes more complex as a video is not merely a sequence of static images. It is a dynamic, multi-dimensional information carrier.
Existing INR-based methods \cite{chen2021nerv,li2022nerv,chen2023hnerv,kwan2024hinerv,kwan2024nvrc,lee2023ffnerv} often represent videos from a superficial perspective, failing to capture the intricate structures within video data. To enhance video reconstruction quality, it is crucial to adopt a comprehensive approach that captures the multi-dimensional nature of videos and optimizes feature utilization in the network architecture design.

In this paper, we propose MSNeRV, an efficient neural video representation with multi-scale feature fusion. Our framework improves network expressiveness from perspectives of temporal, spatial, and network architecture. 
Specifically, in the temporal domain, adjacent frames exhibit strong correlations, and video backgrounds tend to remain stable over short time spans. Leveraging these properties, at the encoding stage, we fuse adjacent features in the time domain and introduce GoP-level  grids to represent background features in a Group of Pictures (GoP). By integrating multi-scale temporal information from single frames, adjacent frames, and GoP-level frames, MSNeRV mitigates temporal inconsistency in rapidly changing videos and enables efficient bitrate allocation across frames.
In the spatial domain, INR-based methods typically rely on continuous upsampling for decoding, where parameter information is incorporated into the video throughout this process. 
Existing methods primarily focus on the final reconstructed video while overlooking intermediate features.
However, we observe that these low-resolution features also contain rich semantic information. 
This omission limits the utilization of spatial information and ultimately degrades reconstruction quality in complex scenes. To address this, we employ multi-resolution supervision to guide video decoding, and introduce a high-frequency boosting module to enhance detail reconstruction. 

Existing INR-based methods predominantly use simple linear structures, leading to insufficient feature exploitation. In fact, we find that integrating parallel architectures to broaden the network allows for more effective processing of intricate video content. Furthermore, inspired by ResNet \cite{he2016deep}, the complementary nature of features from different depths is leveraged in our model.
By incorporating cross-layer connections, MSNeRV maximizes feature utilization, enhancing both its representational capacity and generalization ability.

In summary, our main contributions are as follows:
\begin{itemize}[leftmargin = 2em]
    \item 
    We propose an INR-based video compression framework that systematically incorporates \textbf{multi-scale structure priors} in both temporal and spatial domains. It addresses the limitations of existing INRs in capturing complex motion patterns and high-frequency details.
    \item 
    For MSNeRV, we conduct a multi-scale architecture with two components:
    1) a \textbf{temporal encoder} integrating adjacent-frame feature fusion and GoP-level background modeling, 2) a \textbf{spatial decoder} combining multi-resolution supervision and high-frequency boosting.
    The network is further strengthened by \textbf{multi-scale feature blocks} that maximize feature reuse across layers.
    \item 
    Extensive experiments on the UVG \cite{mercat2020uvg} and HEVC ClassB \cite{sullivan2012overview} datasets demonstrate that MSNeRV outperforms existing INR-based methods in representation capacity, and achieves superior compression performance to VTM on average.
\end{itemize}

%% file: epoch.tex
\begin{figure}[tp]
    \centering
    \includegraphics[width=\linewidth]{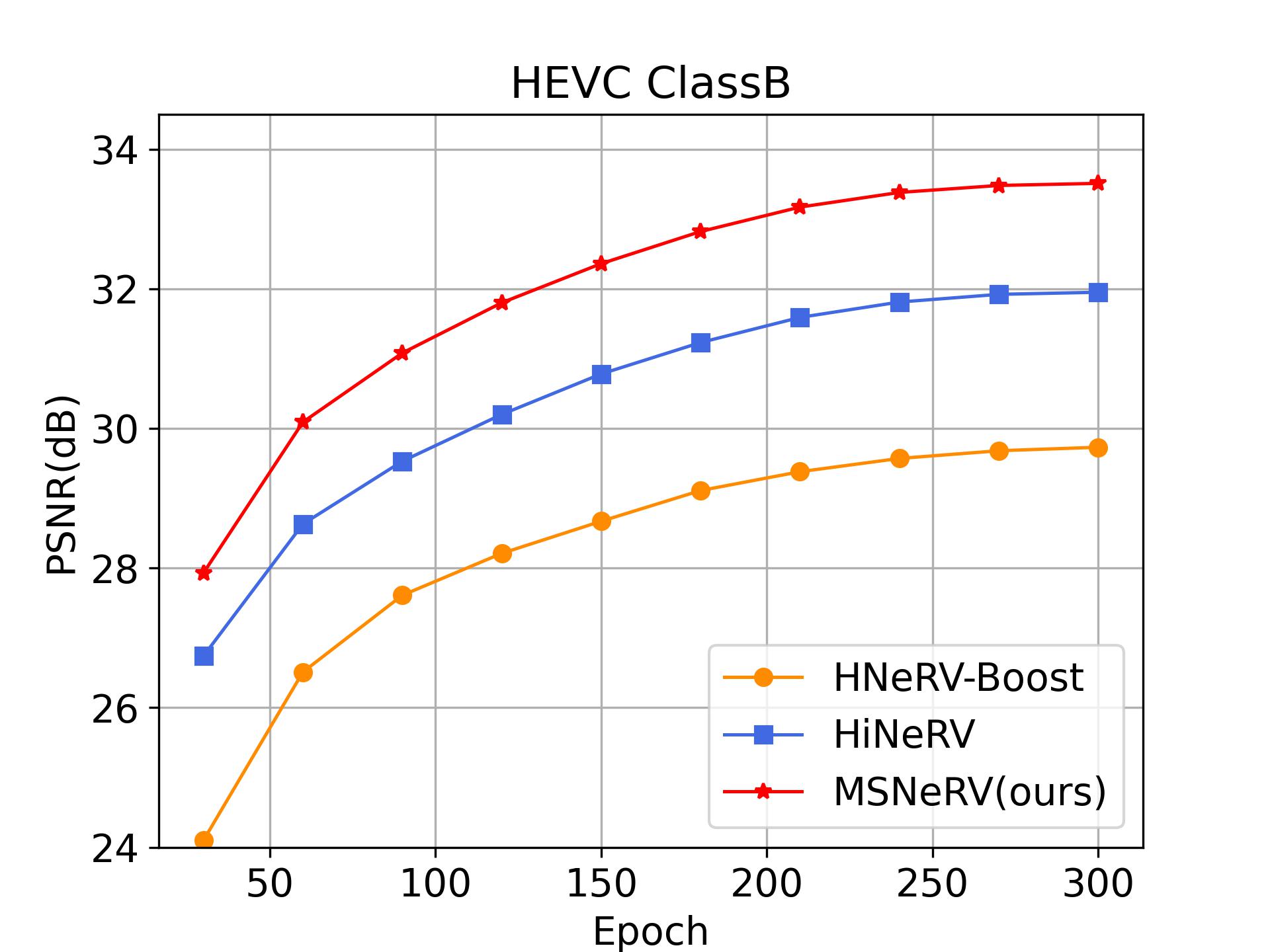}
    \caption{Reconstruction quality over iterations for implicit neural video representations.}
    \label{fig:epoch}
\end{figure}

%% file: 02_related.tex
\section{Related Work}
\label{sec:related}

\subsection{Video Compression}
Video compression is a cornerstone of compute vision and modern multimedia technology. 
The widely used H.264 \cite{wiegand2003overview} and HEVC \cite{sullivan2012overview} standards are generally based on a hybrid framework that integrates prediction, transformation, quantization and entropy encoding. 
A key component of this framework is the Group of Pictures (GoP) structure, which organizes frames in a predefined sequence. 
It typically starts with an I-frame (Intra-coded frame), followed by a mix of P-frames (Predictive frames) and B-frames (Bidirectional frames).
Spatial and temporal predictions (i.e., intra-frame and inter-frame prediction) are employed in different kinds of frames for redundancy reduction.
Intra-frame prediction is primarily used in I-frames. It exploits the spatial correlation within a video by predicting pixel values based on already encoded neighboring pixels within the same frame. 
In contrast, inter-prediction is primarily used in P-frames and B-frames. It reduces temporal redundant information by predicting blocks from previous or future frames.
The latest video compression standard H.266/VVC \cite{bross2021overview}, also known as Versatile Video Coding, achieves $\sim$50\% bitrate reduction compared to H.265/HEVC \cite{sullivan2012overview} at the same visual quality. Advanced block partitioning \cite{huang2021block}, improved intra/inter predictions \cite{dong2021fast,zhao2019wide}, and enhanced coding techniques like deep-learning \cite{park2020fast,zhu2020deep} are introduced in H.266/VVC.

Unlike traditional video codecs that rely on predefined mathematical models, learning-based video compression utilizes neural networks to learn optimal compression strategies from video data.
A common approach is to replace certain modules in traditional video coding, such as inter-frame/intra-frame predictions \cite{laude2016deep,chen2017deepcoder,liu2021deep}, entropy coding \cite{minnen2020channel} and loop-filter \cite{kuanar2018deep, zhang2019recursive}.
DVC \cite{lu2019dvc} introduce the first end-to-end video compression framework, which is optimized by rate-distortion(RD) loss.
The DCVC series \cite{li2021deep,li2023neural,li2024neural} adopt a conditional coding framework that utilizes context from feature domains to improve compression performance.
Furthermore, the DCVC-DC \cite{li2023neural} model enhances context diversity by incorporating both temporal and spatial dimensions.

\subsection{INR-based Video Compression}
Implicit neural representation (INR) \cite{sitzmann2020implicit} is a novel approach in the filed of data representation. Continuous signals, such as images \cite{dupont2021coin,kim2024c3,strumpler2022implicit}, videos \cite{chen2021nerv,li2022nerv,Tang_2023_ICCV,chen2023hnerv,kwan2024hinerv}, 3D scenes \cite{mildenhall2021nerf,pumarola2021d,song2023nerfplayer} and other spatiotemporal data \cite{su2022inras,szatkowski2022hypersound,shen2022nerp}, are encoded using neural networks.
INRs parameterize a signal as a function that maps coordinates to corresponding values.
Neural representation for videos (NeRV) \cite{chen2021nerv} encodes videos by training a neural network to output the entire frame when given a frame index. This approach transforms video compression into a model compression task, where the video content is encapsulated within the network's weights.
The following contributions further explore content-adaptive structure \cite{chen2023hnerv}, conditional decoder \cite{zhang2024boosting} and hierarchical positional encoder \cite{kwan2024hinerv} for efficient video compression and reconstruction.
Additionally, techniques such as quantization and entropy coding are employed to compress the representation itself, further reducing the bitrate \cite{kwan2024nvrc,kwan2024hinerv,zhang2024boosting}. 

Aforementioned approaches primarily focus on feature extraction and compression efficiency while overlooking the structural characteristics of video.
While they achieve impressive compression rates in certain scenarios, they tend to struggle with capturing intricate spatial and temporal dependencies, particularly in videos with high dynamics and fine details. Moreover, many of these methods follow a linear progression with sparse feature interactions across layers, limiting their ability to handle complex video scenes.

%% file: 03_method.tex
\section{Method}
\label{sec:method}
In this section, we provide an overview of the proposed MSNeRV. 
As shown in Figure \ref{fig:MSD}, the encoder first generates grids corresponding to the input indices (Section \ref{subsec:MTE}). These grids are cascaded upsampled through multi-scale feature blocks (Section \ref{subsec:MFB}) in the decoder to reconstruct the video. Intermediate features from the decoding process are leveraged for multi-scale supervision (Section \ref{subsubsec:MRS}), while the high-frequency residual of the generated video is used to improve visual quality (Section \ref{subsubsec:MFB}). In addition, a scale-adaptive loss function is employed for model training (Section \ref{subsubsec:SALoss}).
\subsection{Multi-Scale Temporal Encoder}
\input{MSD}
\label{subsec:MTE}
In traditional hybrid video coding \cite{wiegand2003overview,sullivan2012overview,bross2021overview}, the encoder is responsible for compressing raw video data into a compact bitstream.
However, the encoder in INR-based methods \cite{chen2021nerv,lee2023ffnerv,kwan2024hinerv} transforms spatial coordinates $(i,j,t)$ into positional embeddings for subsequent encoding.
Inspired by the grid-based encoder proposed by FFNeRV \cite{lee2023ffnerv}, we adopt multi-resolution temporal grids as base positional embeddings in our encoder:
\begin{equation}
    X_{base}^t = \gamma_{base}(t),0<t\leq T,
\end{equation}
where $X_{base}^t$ represents the base grid associated with the frame index $t$, and $\gamma_{base}$ refers to a feature space composed of learnable feature maps.
Linear interpolation over both temporal and spatial dimensions is applied in $\gamma_{base}$, as the features have different spatial and temporal resolutions for lower model size.

The base grid $X_{base}^t$ is better suited for video representation than the Fourier encoding commonly used in INRs for video \cite{chen2021nerv,chen2023hnerv,sitzmann2020implicit}. 
However, this structure is relatively coarse when compared to the temporal relationship in real-world videos.
As shown in Figure \ref{flow}, motion between frames may follow multiple patterns.
We employ a temporal window with learnable parameters to model these diverse motions.
As depicted in Figure \ref{fig:MTE}, the base grids within each window are fused using these parameters.
By apply a sliding window mechanism, we obtain a sequence of temporal grids, denoted as $X_{temporal}$.
Adjacent frames have overlapping windows, since the content in a video typically transitions gradually.
To ensure computational consistency in temporal boundaries, the base grid size is expanded to $T+l-1$. 
The temporal grids with a window size $l$ can be written as 
\begin{equation}
    X_{temporal}^t=\sum_{i=t}^{i=t+l-1}(w_i^t\cdot X_{base}^i),0<t\leq T .
\end{equation}

Over time, a scene transition may take place in videos, leading to a shift in the background.
To account for this, we divide the video into $K$ Group of Pictures (GoPs). A GoP-level grid, representing the background features, is then added to the temporal grids:
\begin{equation}
    X_{fused}^{(t)}=X_{temporal}^{(t)}+\gamma_{GoP}(k),0<t\leq T, 0<k\leq K,
\end{equation}
where $k$ denotes the index of the \text{k-th} GoP that includes the time step $t$, and $\gamma_{Gop}$ represents the feature space formed by GoP-level grids.

As shown in Figure \ref{fig:MTE}, the grid information corresponding to single frames, adjacent frames and GoP-level frames is integrated using the methods described above. MSNeRV fuses multi-scale temporal features during the encoding process and the final fused grids can be written as:
\begin{equation}
    \begin{split}
        X_{fused}^{t}=\sum_{i=t}^{i=t+l-1}(w_i^t\cdot \gamma_{base}.(t))+\gamma_{GoP}(k),\\
        0<t\leq T, 0<k\leq K.
    \end{split}
\end{equation}

Additionally, we use the unified frame-wise \cite{chen2021nerv,chen2023hnerv} and patch-wise \cite{mentzer2019practical} feature representation, as introduced in HiNeRV \cite{kwan2024hinerv}, to improve training efficiency.

\input{MTE}
\subsection{Multi-Scale Spatial Decoder}
\label{subsec:MSD}
The decoder progressively upsamples the low-resolution encoded grid $X_{fused}\in\mathbb{R}^{T\times h\times w\times C_0}$ to reconstruct the video $X_{N}\in \mathbb{R}^{T\times H\times W\times 3}$.
In this section, we provide a detailed introduction of our multi-scale spatial decoder, focusing on three key innovations: multi-resolution supervision (Section \ref{subsubsec:MRS}), high-frequency boosting (Section \ref{subsubsec:MFB}) and the scale-adaptive loss function (Section \ref{subsubsec:SALoss}).

\subsubsection{Multi-Resolution Supervision}
\label{subsubsec:MRS}
During the training phase, most INR-based methods optimize the network by minimizing the loss between the reconstructive video and the target video.
However, the intermediate features generated during the upsampling process are overlooked.
Visualization of these features, as shown in Figure \ref{fig:six_images}, reveal that these low-resolution features also contain extensive contexture information, such as motion trends and background structures.
They are effective in enhancing the quality of reconstructive videos.

A multi-resolution supervision (MRS) is introduced in our MSNeRV to achieve a coarse-to-fine representation learning.
The resolution level is denoted as $r$, ranging from $1$ to $N$, where $r=1$ represents the lowest resolution and $r=N$ corresponds to the original resolution.
We apply a 3x3 convolution layer $\mathcal{N}_r$ to extract the low-level video representation $\hat{X}_r\in\mathbb{R}^{T\times h_r\times w_r \times 3}$ from the intermediate feature $X_{r}\in\mathbb{R}^{T\times h_r\times w_r\times C_r}$.
$\mathcal{N}_r$ is solely used during the training parse and is not involved in the decoding process. Therefore, it does not introduce any additional bitrate.
Meanwhile, the low-resolution video $V_{r\downarrow }$ is obtained through cascaded downsampling. 
By explicitly supervising intermediate features, MRS encourages structured learning, enhancing both training efficiency and feature representation.
Moreover, MRS ensures that each layer focuses on the most critical aspects of the video at its respective resolution, which leads to a rational and balanced bitstream allocation.
We employ max pooling \cite{lecun1998gradient} for downsampling because it preserves key information such as edges, texture and significant patterns. Additionally, max pooling is invariant to small translations, improving the stability of the patch-wise training.

\subsubsection{High-Frequency Boosting}
\label{subsubsec:MFB}
In addition to multi-resolution features, videos also contain information at different frequency levels.
Most INR-based compression models prioritize fitting the low-frequency components of the video, as they exhibit smoother patterns and account for the majority of the signal's energy \cite{jahne2005digital}.
In contrast, high-frequency components, as shown in Figure \ref{hf}, often involve fine details and rapid changes. They are inherently more complex to model.
Decomposing the video into static and dynamic elements \cite{yan2024ds,kim2024snerv} or incorporating additional high-frequency networks \cite{tang2025canerv} can enhance visual quality of sharp details.

In MSNeRV, we introduce a novel high-frequency boosting strategy. It can achieve similar benefits during optimization, without altering the network structure.
Specifically, we compute the residual between the generated video $X_N$ and the original video $V_{gt}$. Then a high-pass filter $\mathcal{H}$ is applied to extract high-frequency components from this residual.
These high-frequency details are added back to the original video, serving as a reference for subsequent loss calculation:
\begin{equation}
    V_{boost}=V_{gt}+\mathcal{H}(|V_{gt}-X_N|).
\end{equation}

\input{visual}
\subsubsection{Scale-Adaptive Loss}
\label{subsubsec:SALoss}
We adopt a scale-adaptive loss (SA Loss) as the loss function for MSNeRV. It combines the L1, MSE and MS-SSIM losses:
\begin{equation}
    \mathcal{L}_{SA}^{(r)} = \alpha_r \mathcal{L}_{MSE}^{(r)}+\beta_r \mathcal{L}_{1}^{(r)} +(1-\alpha_r -\beta_r)\mathcal{L}_{MS-SSIM}^{(r)},
\end{equation}
where $\alpha_r$ and $\beta_r$ are coefficients determined by the resolution level $r$.
Notably, we ensure that $\alpha_r+\beta_r$ remains $1$ for low-resolution videos, meaning that MS-SSIM is exclusively applied at original resolution. Since this metric primarily focuses on the perceptual similarity that is crucial for the final reconstructed video.

Given the formulas of $\mathcal{L}_{MSE}$ and $\mathcal{L}_1$:  
\begin{equation}
\begin{aligned}
    \mathcal{L}_{MSE}^{(r)}=\frac{1}{T}\sum_{t=1}^T||\mathcal{N}_r(X_{r}^{(t)})-V_{r\downarrow }||_2^2,\\
    \mathcal{L}_{1}^{(r)}=\frac{1}{T}\sum_{t=1}^T||\mathcal{N}_r(X_{r}^{(t)})-V_{r\downarrow }||_1,
\end{aligned}
\end{equation}
we observe that $\mathcal{L}_{MSE}$ is more sensitive to large errors, while $\mathcal{L}_1$ is more robust with noisy data.
Prior studies \cite{zhao2016loss,janocha2017loss} also indicate that $\mathcal{L}_{MSE}$ tends to produce smooth but blurry results, whereas $\mathcal{L}_1$ better retains edges and textures.
Based on discussions above, MSNeRV increases $\alpha_r$ at lower resolutions, while decreasing $\beta_r$.

The total loss in MSNeRV is the sum of losses across all scales, with the high-frequency enhanced video serving as the reference at the original resolution:
\begin{equation}
    \mathcal{L}_{total}=\sum_{r=1}^{N-1}\mathcal{L}_{SA}^{(r)}(V_{r\downarrow},\hat X_r)+\mathcal{L}_{SA}^{(N)}{(V_{boost},X_N)}.
\end{equation}
By incorporating both multi-resolution and multi-frequency information into the loss function, MSNeRV achieves higher fidelity reconstructions and preserves fine details.

\input{MFB}

\subsection{Multi-Scale Feature Block}
\label{subsec:MFB}
In MSNeRV, feature upsampling and detail restoration are performed through Multi-Scale Feature blocks (MSF blocks).
Firstly, the input feature $X_{n-1}$ with size $T\times h_{n-1}\times w_{n-1}\times C_{n-1}$ is processed by a hybrid upsample layer to obtain the high-resolution feature $\tilde X_{n-1\uparrow}\in\mathbb{R}^{T\times h_n\times w_n\times C_{n-1}}$.
This hybrid upsample layer combines bilinear interpolation $U_b$ and parallel pixel shuffle \cite{shi2016real} $U_p^n$ to  enhance detail preservation.
A hierarchical encoding $\gamma_n(i,j,t)$, computed using the local grid introduced in HiNeRV \cite{kwan2024hinerv}, is added to $\tilde X_{n-1\uparrow}$ to enhance the encoder's capacity:
\begin{equation}
    X_{n-1\uparrow}=U_b(X_{n-1})+U_P^n(X_{n-1})+\gamma_n(i,j,t),
\end{equation}
where $X_{n-1\uparrow}$ denotes the final upsampled feature, and $(i, j, t)$ refers to the patch coordinate.

Subsequently, $K$ fusion layers are applied to further enhance the feature representation:
\begin{equation}
    X_n^{(K) }= \mathcal{F}_K(...\mathcal{F}_2(\mathcal{F}_1(X_{n-1\uparrow})),
\end{equation}
where $X_n^{(k)}$ represents the output feature of the $k$-th fusion layer $\mathcal{F}_k$.
$\mathcal{F}_{k}$ fuses features from depth-wise convolutions with different kernel sizes to get the fused feature $\hat{X}_n^{(k)}$, and employs an MLP layer for channel projection.
As shown in Figure \ref{33}, \ref{55}, and \ref{mlp}, the features processed by different layers exhibit diverse visualization patterns.
To efficiently utilize these features, we fuse $\hat{X}_n^{(k)}$ with the MLP-processed feature by performing a channel slicing.

Additionally, to fully leverage feature information at different depth, we employ a cross-depth fusion layer $\mathcal{G}_n$ to integrate all these features into $X_n^{(K)}$.
The output feature $X_n$ with a size of  $T\times h_n\times w_n\times C_n$ can be written as:
\begin{equation}
    X_n=X_n^{(K)}+\mathcal{G}_n(X_n^{(1)}\oplus\cdots\oplus X_n^{(K-1)} \oplus X_n^{(K)}),
\end{equation}
where ``$\oplus$'' represents the concatenation along the channel dimension.
The contributions of these components are analyzed through ablation studies in Section \ref{subsec:ablation}.

%% file: MSD.tex
\begin{figure*}[tp]
    \centering
    \includegraphics[width=\linewidth]{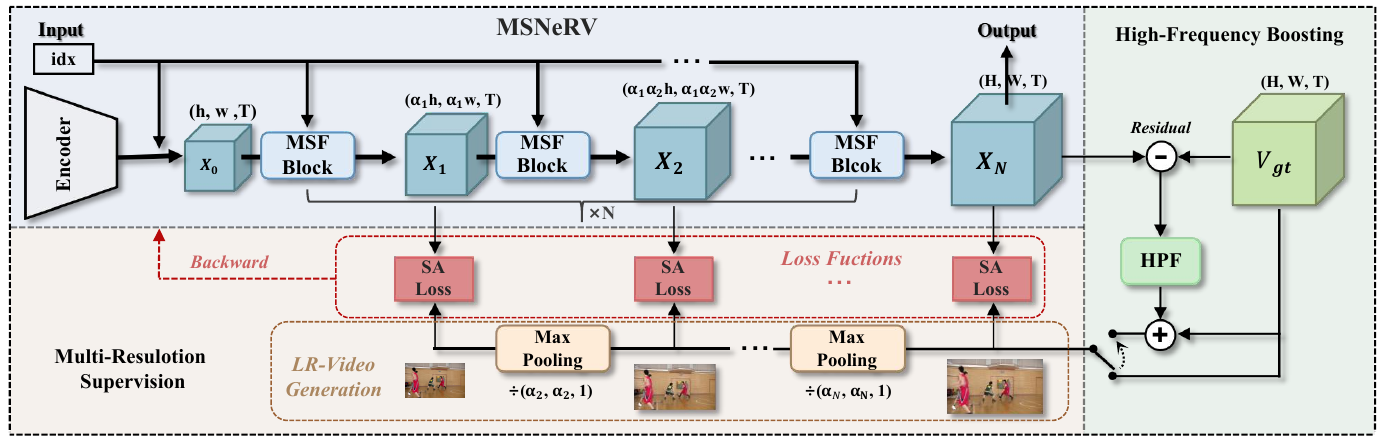}
    \caption{
    \textbf{Left top}: Main architecture of MSNeRV. 
    \textbf{Right}: High-frequency boosting. The residual between $X_N$ and $V_{gt}$ is processed by the high pass filter (HPF) and selectively added to $V_{gt}$.
   \textbf{Left bottom}: Multi-resolution supervision. Low-resolution videos are generated by cascaded max-pooling. Scale-adaptive loss (SA Loss) functions are used to incorporate multi-scale spatial information into the network.}
    \label{fig:MSD}
\end{figure*}

%% file: MTE.tex
\begin{figure}[tp]
    \centering
    \includegraphics[width=\linewidth]{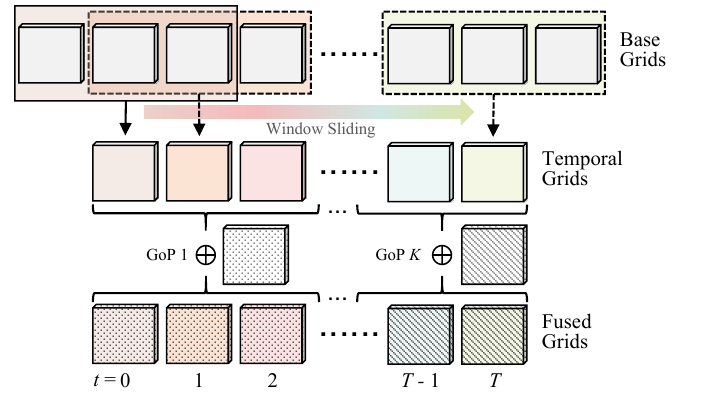}
    \caption{The multi-scale temporal encoder. 
    We use a temporal window to capture the variations between adjacent frames, and then integrate a  GoP-level grid to get the fused grids. }
    \label{fig:MTE}
\end{figure}

%% file: visual.tex
\begin{figure}[tp]
    \centering
    \begin{subfigure}{0.3\columnwidth}
        \centering
        \includegraphics[width=\linewidth]{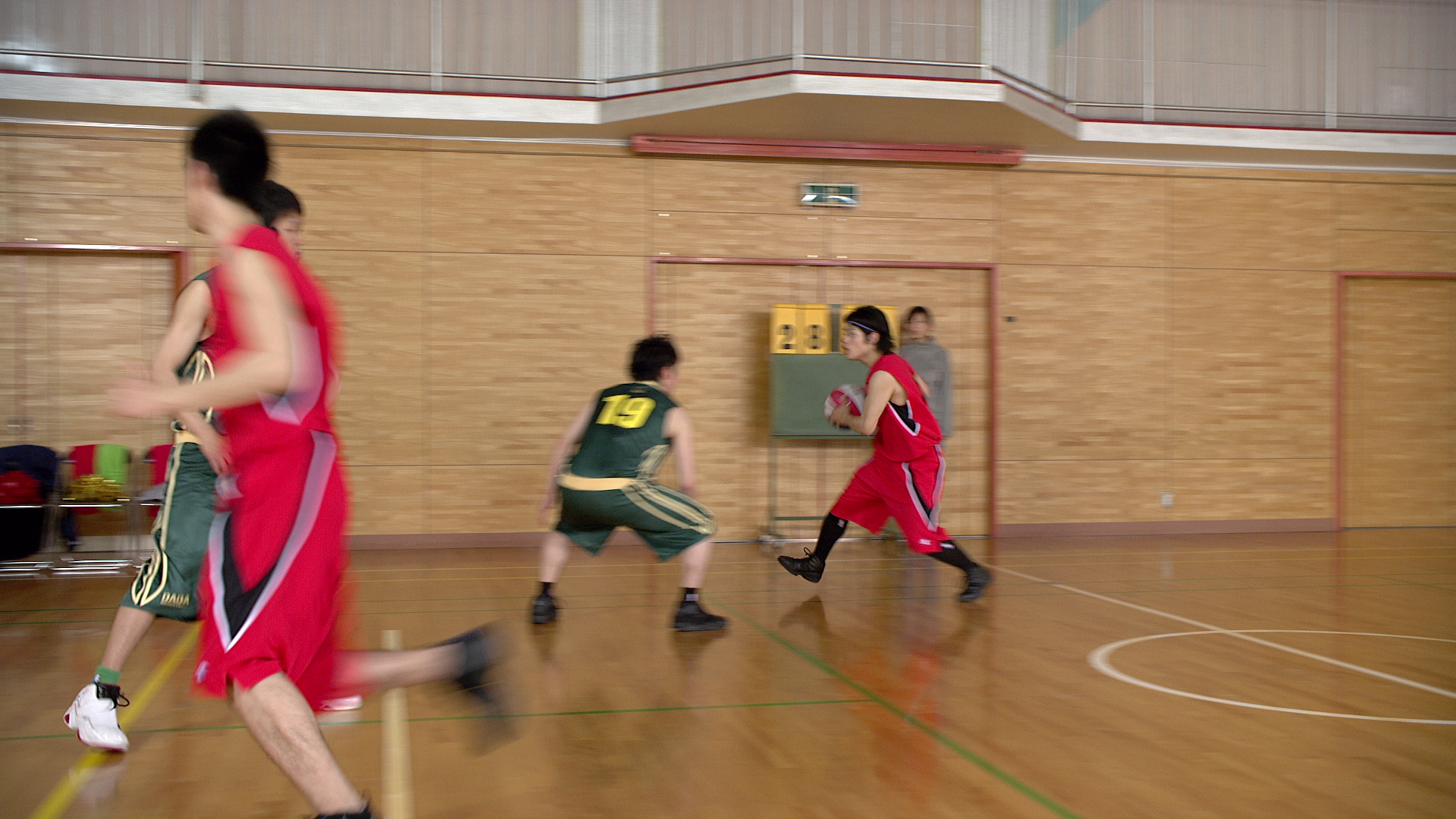}
        \caption{Ground Truth}
    \end{subfigure}
    \begin{subfigure}{0.3\columnwidth}
        \centering
        \includegraphics[width=\linewidth]{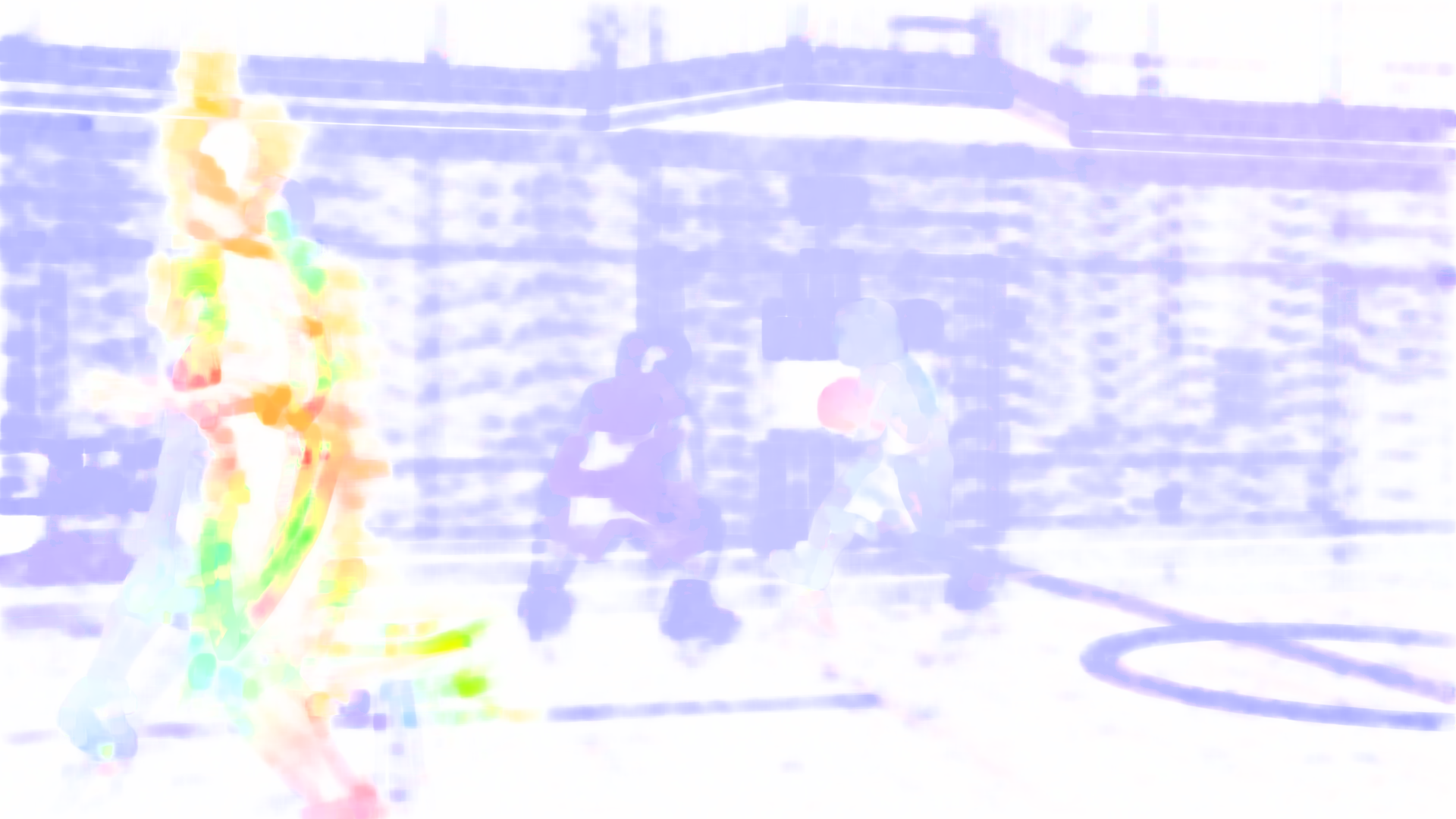}
        \caption{Optical flow}
        \label{flow}
    \end{subfigure}
    \begin{subfigure}{0.3\columnwidth}
        \centering
        \includegraphics[width=\linewidth]{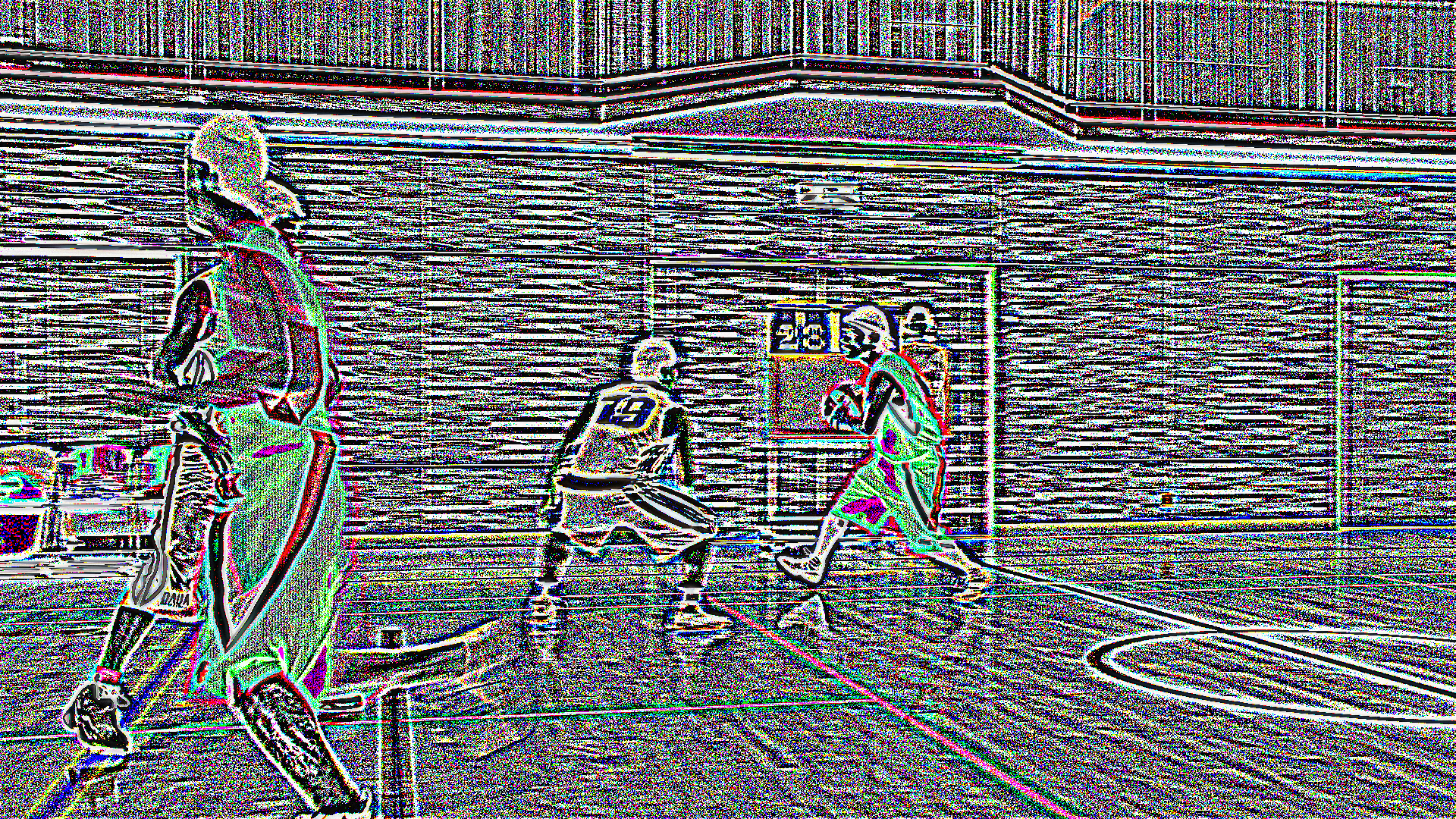}
        \caption{High Frequency}
        \label{hf}
    \end{subfigure}

    \begin{subfigure}{0.3\columnwidth}
        \centering
        \includegraphics[width=\linewidth]{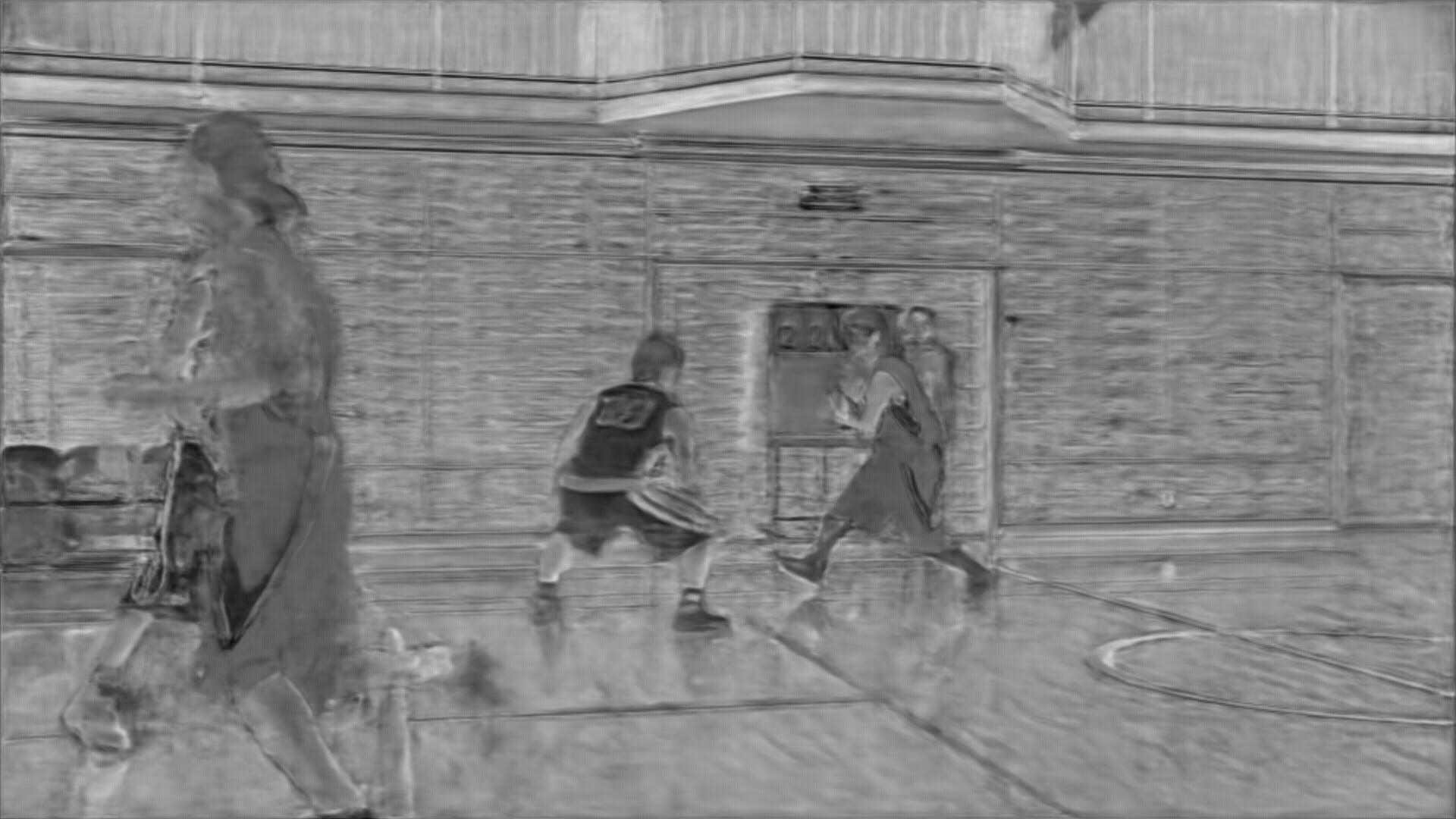}
        \caption{$3\times3$ Conv.}
        \label{33}
    \end{subfigure}
    \begin{subfigure}{0.3\columnwidth}
        \centering
        \includegraphics[width=\linewidth]{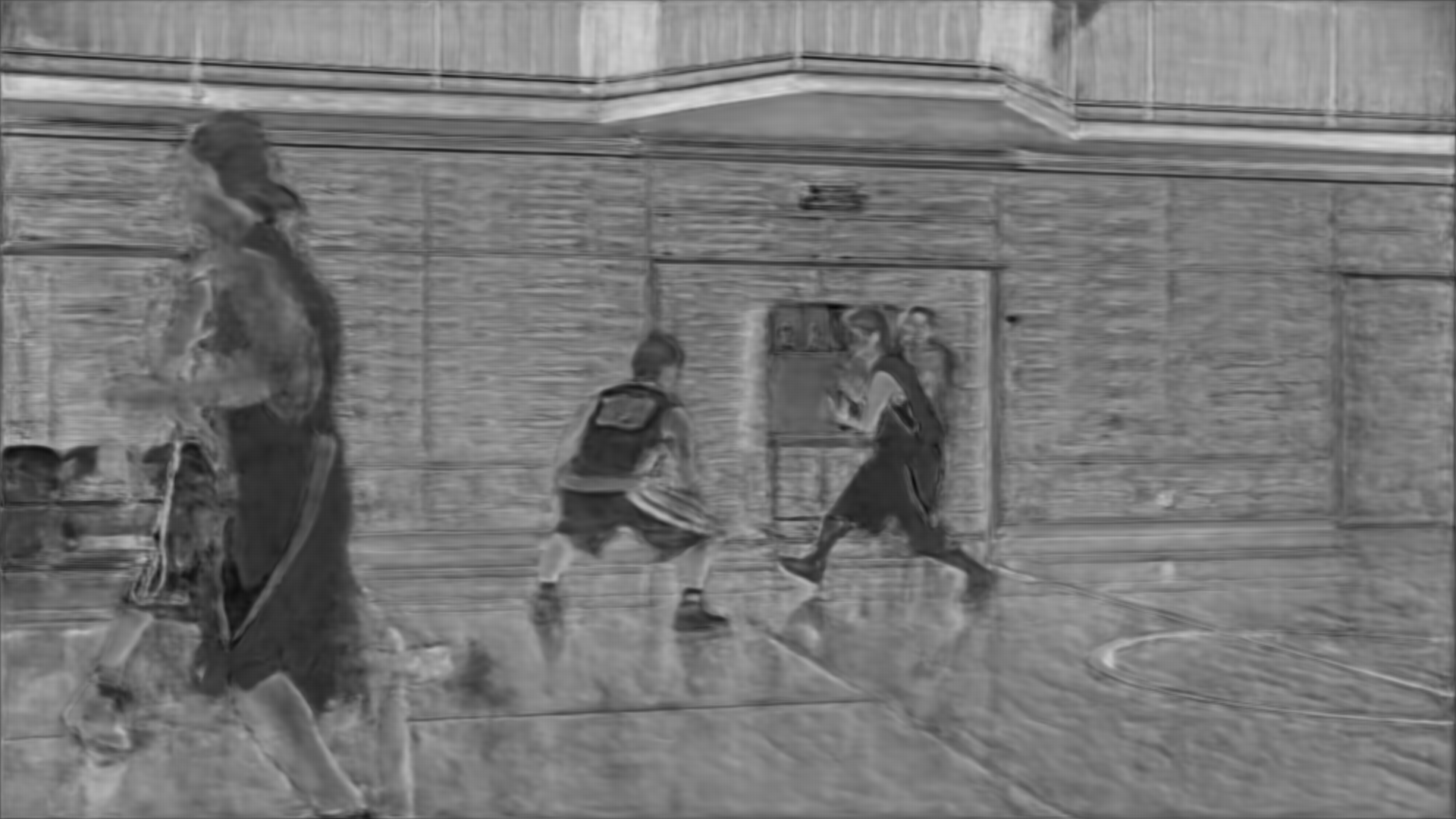}
        \caption{$5\times 5$ Conv.}
        \label{55}
    \end{subfigure}
    \begin{subfigure}{0.3\columnwidth}
        \centering
        \includegraphics[width=\linewidth]{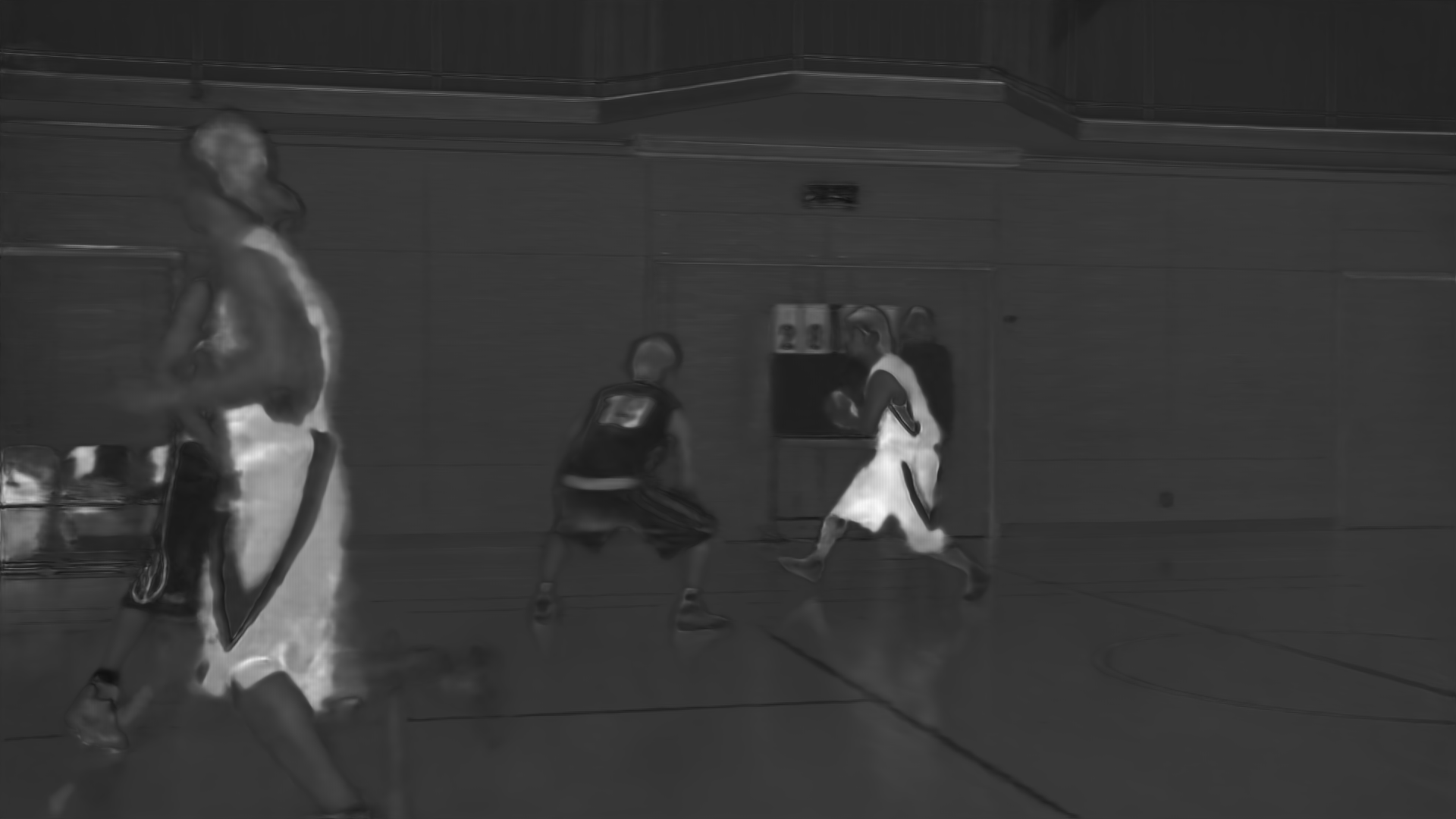}
        \caption{MLP}
        \label{mlp}
    \end{subfigure}

    \caption{\textbf{(a)} Video Ground Truth. \textbf{(b)} Optical flow map where different colors indicate varying motion directions and the white areas represent regions with no motion. \textbf{(c)} High-Frequency components extracted by a high-pass filter. \textbf{(d)(e)(f)} Feature maps extracted by the $3\times3$ convolution, $5\times5$ convolution and MLP layers. }
    \label{fig:six_images}
\end{figure}

%% file: MFB.tex
\begin{figure}[tp]
    \centering
    \includegraphics[width=\linewidth]{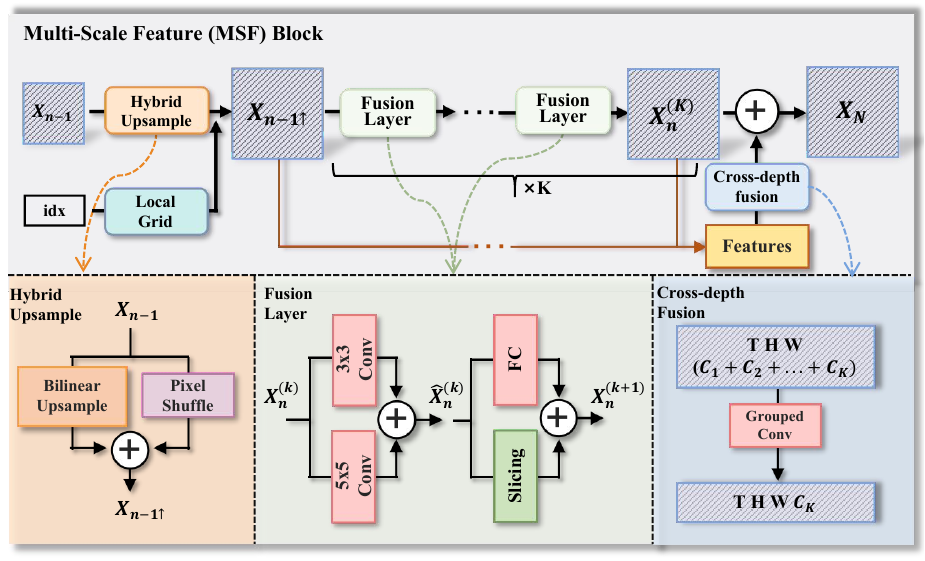}
    \caption{
    \textbf{Up}: The multi-scale feature (MSF) block used in MSNeRV decoder.
    \textbf{Bottom left}: The hybrid upsample.
    \textbf{Bottom middle}: The fusion layer, where features processed by different layers are fused.
    \textbf{Bottom right}: The cross-depth fusion layer, incorporating information at different depths.
    }
    \label{fig:MFB}
\end{figure}

%% file: 04_experiments.tex
\section{Experiments}
\label{sec:experiments}

\input{tables/regress}
\subsection{Video Representation}
We conduct experiments on 12 video sequences with a resolution of 1920$\times$1080, including 5 videos from HEVC ClassB \cite{sullivan2012overview} and 7 videos from the UVG \cite{mercat2020uvg} dataset.
All videos are captured from real-world scenes.
To evaluate the representation capability of our model, we compare it with HiNeRV \cite{kwan2024hinerv} and HNeRV-boost \cite{zhang2024boosting}.
For a fair comparison, we ensure that all model sizes are within a similar range. We use open-source implementations of these methods and adhere to their default settings.
All models are trained for 300 epochs, and the Peak Signal-to-Noise Ratio (PSNR) metric is used to evaluate reconstruction quality.
For MSNeRV, we use the Adam optimizer \cite{kingma2014adam} with an initial learning rate of 2e-3 and adopt the SA-loss as described in Section \ref{subsec:MSD} .

Table \ref{tab:regression} presents the model sizes of all methods alongside representation results across all video sequences.
Existing INR-based methods tend to suffer from noticeable distortions in fast-motion regions and intricate details (e.g., human faces), particularly in videos such as \textit{BasketballDrive} and \textit{BQTerrace}. 
However, our method captures inter-frame relationships through its temporal fusion mechanism, allowing for more effective modeling of dynamic scenes. The multi-scale spatial decoder also enhances the model's ability to reconstruct fine details.
As a result, MSNeRV excels in dynamic sequences, and also performs better on static videos like \textit{HoneyBee} and \textit{Beauty}.
Moreover, MSNeRV maintains consistently better performance throughout the training process, as shown in Figure \ref{fig:epoch}.



\subsection{Video Compression}
\input{tables/compression}
\input{perfomance}
\input{compare}
We compare MSNeRV with traditional codec ( e.g., VTM-23.7 with Random Access configuration \cite{browne2022algorithm}), learning-based approaches ( e.g., DCVC-DC \cite{li2023neural}, DCVC-FM \cite{li2024neural}) and INR-based methods (e.g., HiNeRV, HNeRV-boost). 
The performance of DCVC-DC and DCVC-FM is evaluated using the implementation and pre-trained models provided by the authors.
INR-based methods employ their own strategy to compress network parameters.
Our model is trained for 30 additional epochs of Quantization-Aware Training (QAT) \cite{jacob2018quantization} to minimize quality loss during quantization. We apply the entropy coding \cite{mentzer2019practical} to generate the actual bitstream.
MS-SSIM is used to evaluate the subjective quality of reconstructed videos alongside with PSNR. Compression performance is assessed using bits per pixel (bpp), and Bj\o ntegaard Delta bitrate (BDBR) \cite{bjontegaard2008improvements} is employed to measure the bitrate savings across different codecs.

Table \ref{tab:compression} presents the BD-rate results, reflecting the bitrate saving or increase of our method compared to others when achieving the same video quality. MSNeRV outperforms the state-of-the-art conventional codec, VTM (RA), on the HEVC ClassB \cite{sullivan2012overview} dataset that contains more dynamic video content. This highlights the suitability of MSNeRV for real-world scenes with complex motion.
On the UVG \cite{mercat2020uvg} dataset, MSNeRV also achieves performance close to VTM (RA), demonstrating its effectiveness across a diverse range of video content.
Under similar network coding schemes, MSNeRV achieves a $42\%$ bitrate saving on the HEVC ClassB \cite{sullivan2012overview} dataset and a $28\%$ bitrate reduction on the UVG \cite{mercat2020uvg} dataset compared to HiNeRV. These results confirm that our model achieves the best representation capability in INR-based video coding.

Figure \ref{fig:performance} illustrates Rate-Distortion curves for the UVG \cite{mercat2020uvg}\ and HEVC ClassB \cite{sullivan2012overview} datasets.
Our method achieves superb performance across different bitrates.
In addition, MSNeRV outperforms all other approaches in terms of MS-SSIM, indicating a higher subjective quality of reconstructive videos.
The visualization results in Figure \ref{fig:compare} further illustrate that MSNeRV achieves more effective motion modeling compared to HiNeRV.
Moreover, both MSNeRV and HiNeRV outperform VTM in compression regions with structured patterns, such as text.
This is because such regions can be represented by relatively simple functions, making them more suitable for INR-based modeling.

\subsection{Ablation Studies}
\input{tables/ablation}
\input{ent}
To further evaluate the effectiveness of our proposed method, we conduct ablation studies on the HEVC ClassB \cite{sullivan2012overview} dataset by systematically removing or modifying certain components. Specifically, we compare our model against the following variants: \textbf{V1} without the temporal fusion mechanism and GoP-level grids, \textbf{V2} without multi-resolution supervision, \textbf{V3} without high-frequency enhancement, \textbf{V4} replacing the hybrid upsampling layer with trilinear upsampling, \textbf{V5} replacing the hybrid upsampling layer with pixel shuffle, \textbf{V6} replacing the fusion layer with a $3\times3$ convolution followed by an MLP layer, \textbf{V7} without cross-depth fusion.

As shown in Table \ref{tab:ablation}, MSNeRV achieves the best RD performance.
In \textbf{V1}, \textbf{V2} and \textbf{V3}, although the model size remains nearly unchanged, the BPP increase. This suggests that incorporating multi-scale information in either spatial or temporal domain leads to a more efficient allocation of network parameters, as illustrated in Figure \ref{fig:ent}.
The results for \textbf{V4} and \textbf{V5} demonstrate the effectiveness of the hybrid upsampling approach compared to individual methods.
It can be indicated from the results of \textbf{V6} and \textbf{V7} that the model's expressive power is significantly enhanced with feature fusion across different layers.
These fusion strategies allow the model to capture more intricate and diverse patterns in the data, with little additional parameters.
\label{subsec:ablation}

%% file: tables/regress.tex
\begin{table*}[t]
\centering
 \resizebox{\textwidth}{!}{%
\begin{tabular}{c c|| c c c c c || c c c c c c c || c}
\specialrule{2pt}{0pt}{2pt}
\multirow{2}{*}{Model}&\multirow{2}{*}{Size} &\multicolumn{5}{c||}{HEVC ClassB}&\multicolumn{7}{c||}{UVG}& \multirow{2}{*}{avg.}\\
 \cmidrule(lr){3-7} \cmidrule(lr){8-14}
 && Bas. & BQT. & Cac. & Kim. & Par. & Bea. & Bos. & Hon. & Joc. & Rea. & Sha. & Yac. \\
\specialrule{1pt}{2pt}{2pt}
\specialrule{1pt}{0pt}{2pt}
\thead{HNeRV-Boost\\}&3.05M &28.49&27.79&30.09&32.55&29.71&33.70&35.52&39.50&33.62&27.60 &35.50&29.10&31.93\\
\thead{HiNeRV}&3.15M  &31.43&31.19&31.46&34.41&31.25&34.07&38.58&39.67&36.12&31.44&35.75&30.89&33.86\\
\thead{MSNeRV}&3.03M &\textbf{32.68}&\textbf{32.27}&\textbf{32.70}&\textbf{35.80}&\textbf{32.75}&\textbf{34.17}&\textbf{38.95}&\textbf{39.71}&\textbf{37.37}&\textbf{32.59}&\textbf{35.85}&\textbf{31.55}&\textbf{34.70}\\
\specialrule{1pt}{2pt}{2pt}
\thead{HNeRV-Boost}&6.62M &31.17&30.31&32.71&35.09&32.04&34.18&38.28&39.75&36.02&31.05&36.60&31.35&34.05\\
\thead{HiNeRV}&6.42M& 33.45&32.54&33.14&36.41&32.86&34.31&40.29&39.79&37.87&34.45&36.90&32.85&35.41\\
\thead{MSNeRV}&5.92M& \textbf{34.30}&\textbf{33.16}&\textbf{34.09}&\textbf{37.34}&\textbf{34.50}&\textbf{34.41}&\textbf{40.50}&\textbf{39.80}&\textbf{38.61}&\textbf{35.26}&\textbf{36.93}&\textbf{33.42}&\textbf{36.03}\\

\specialrule{1pt}{2pt}{2pt}
\specialrule{1pt}{0pt}{0pt}
\end{tabular}

 }
\caption{
Video regression results (PSNR$\uparrow$) on HEVC ClassB \cite{sullivan2012overview} and UVG \cite{mercat2020uvg} datasets for two ranges of scales.
}
\vspace{-0.05in}
\label{tab:regression}
\end{table*}

%% file: tables/compression.tex
\begin{table*}[t]
\centering
 \resizebox{0.8\textwidth}{!}{%
\begin{tabular}{c c c c c c c}
\specialrule{1.5pt}{0pt}{2pt}
Dataset&Metric&HiNeRV&HNeRV-Boost&DCVC-DC&DCVC-FM&VTM(RA)\\
\specialrule{1pt}{0pt}{2pt}
\multirow{2}{*}{ClassB}&PSNR&$-41.9\%$& $-73.4\%$& $-22.6\%$&$-25.9\%$&$-4.5\%$\\
                    &MS-SSIM&$-27.5\%$&$-86.4\%$ & $-43.8\%$&$-45.5\%$&$-30.5\%$\\
\specialrule{1pt}{0pt}{2pt}
\multirow{2}{*}{UVG}&PSNR&$-26.8\%$&$-65.4\%$ &$-6.3\%$&$-10.8\%$&$7.8\%$\\
                    &MS-SSIM&$-17.4\%$&$-75.0\%$ &$-29.8\%$&$-32.9\%$&$-26.2\%$\\
\specialrule{1.5pt}{0pt}{2pt}

\end{tabular}

 }
\caption{
BD-Rate (Measured in PSNR/MS-SSIM) results on HEVC ClassB \cite{sullivan2012overview} and UVG \cite{mercat2020uvg} datasets.
}
\vspace{-0.05in}
\label{tab:compression}
\end{table*}

%% file: perfomance.tex
\begin{figure*}[tp]
    \centering
    \begin{subfigure}{0.4\textwidth}
        \centering
        \includegraphics[width=\linewidth]{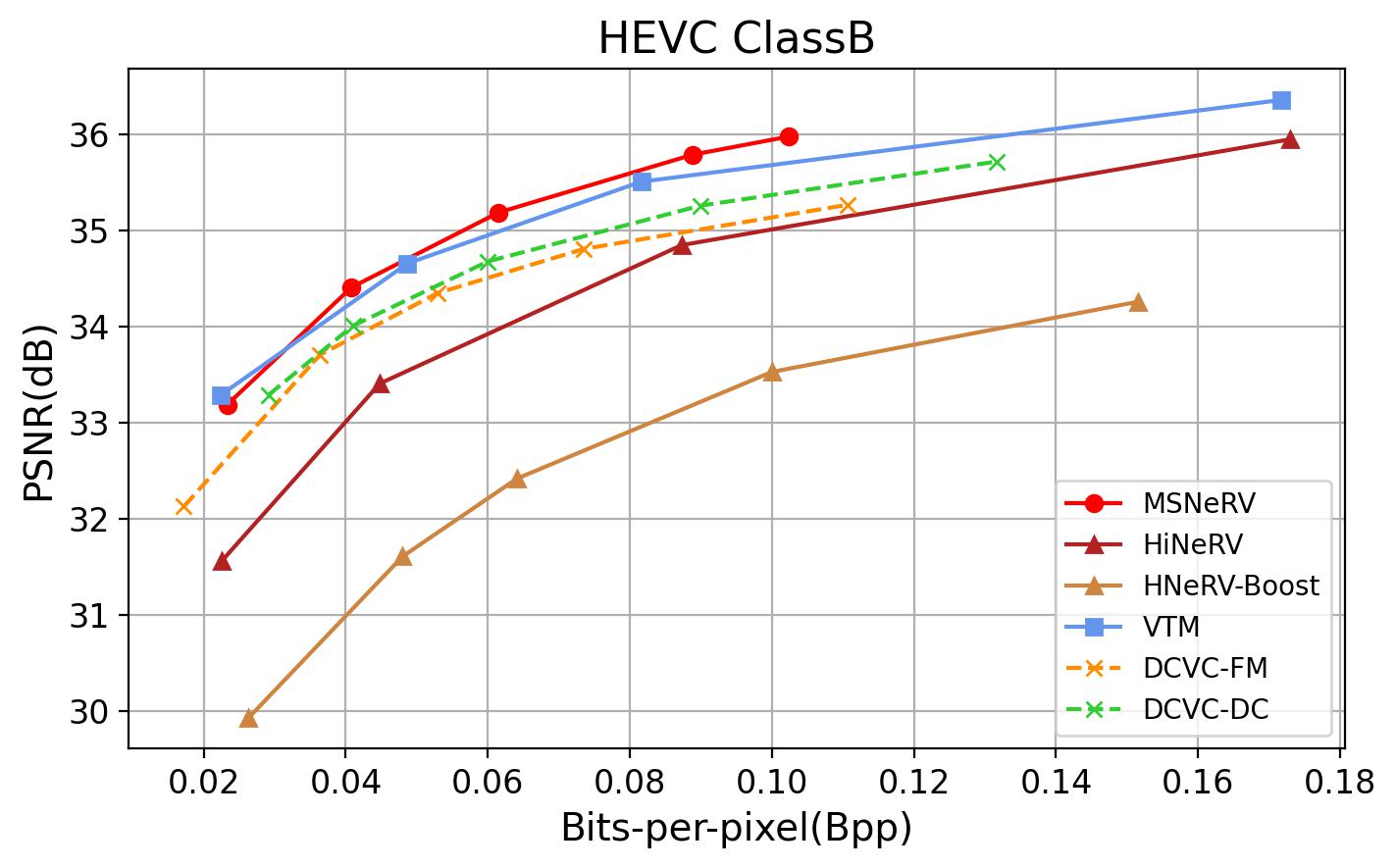}
    \end{subfigure}
    \begin{subfigure}{0.4\textwidth}
        \centering
        \includegraphics[width=\linewidth]{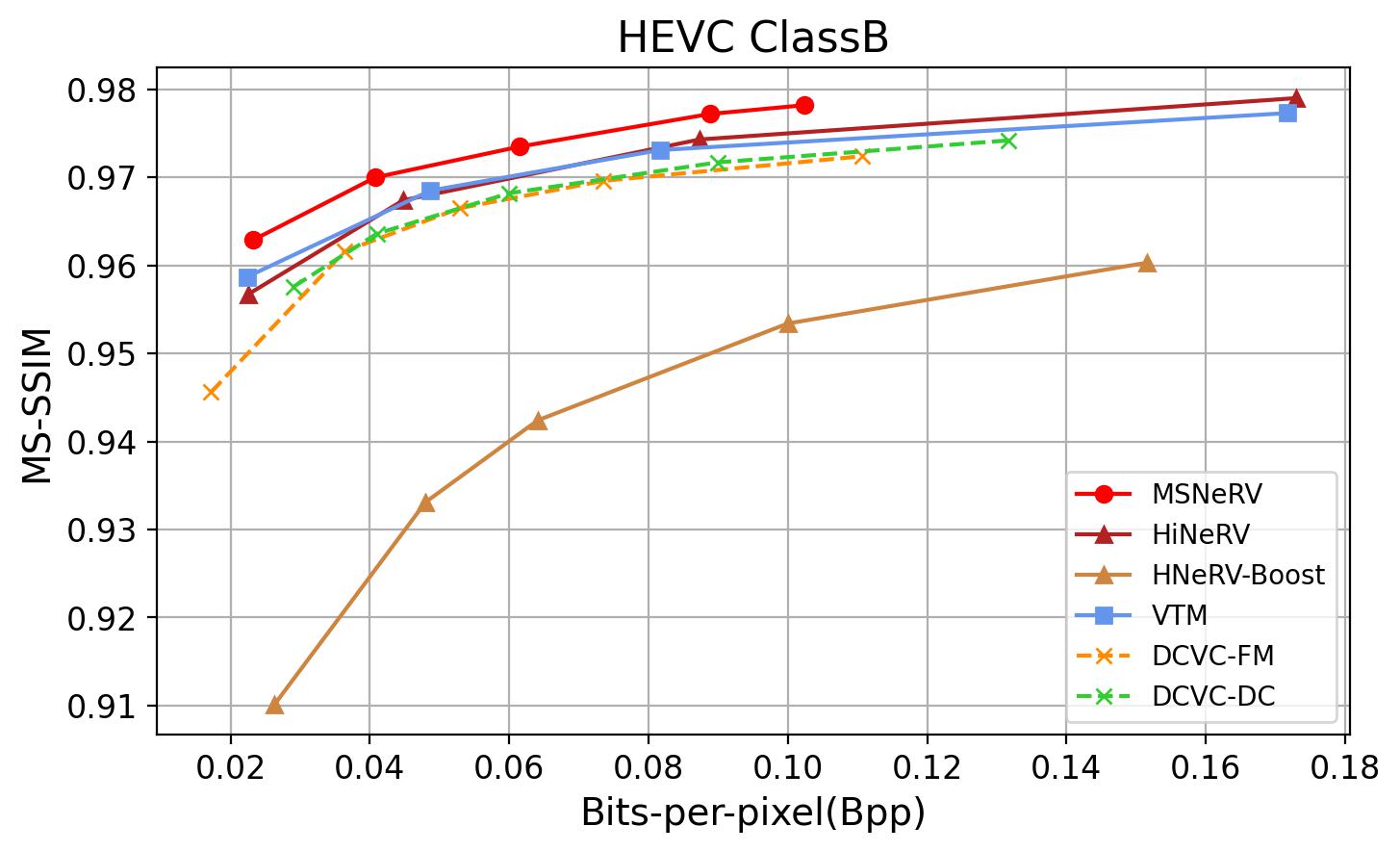}
    \end{subfigure}

    \begin{subfigure}{0.4\textwidth}
        \centering
        \includegraphics[width=\linewidth]{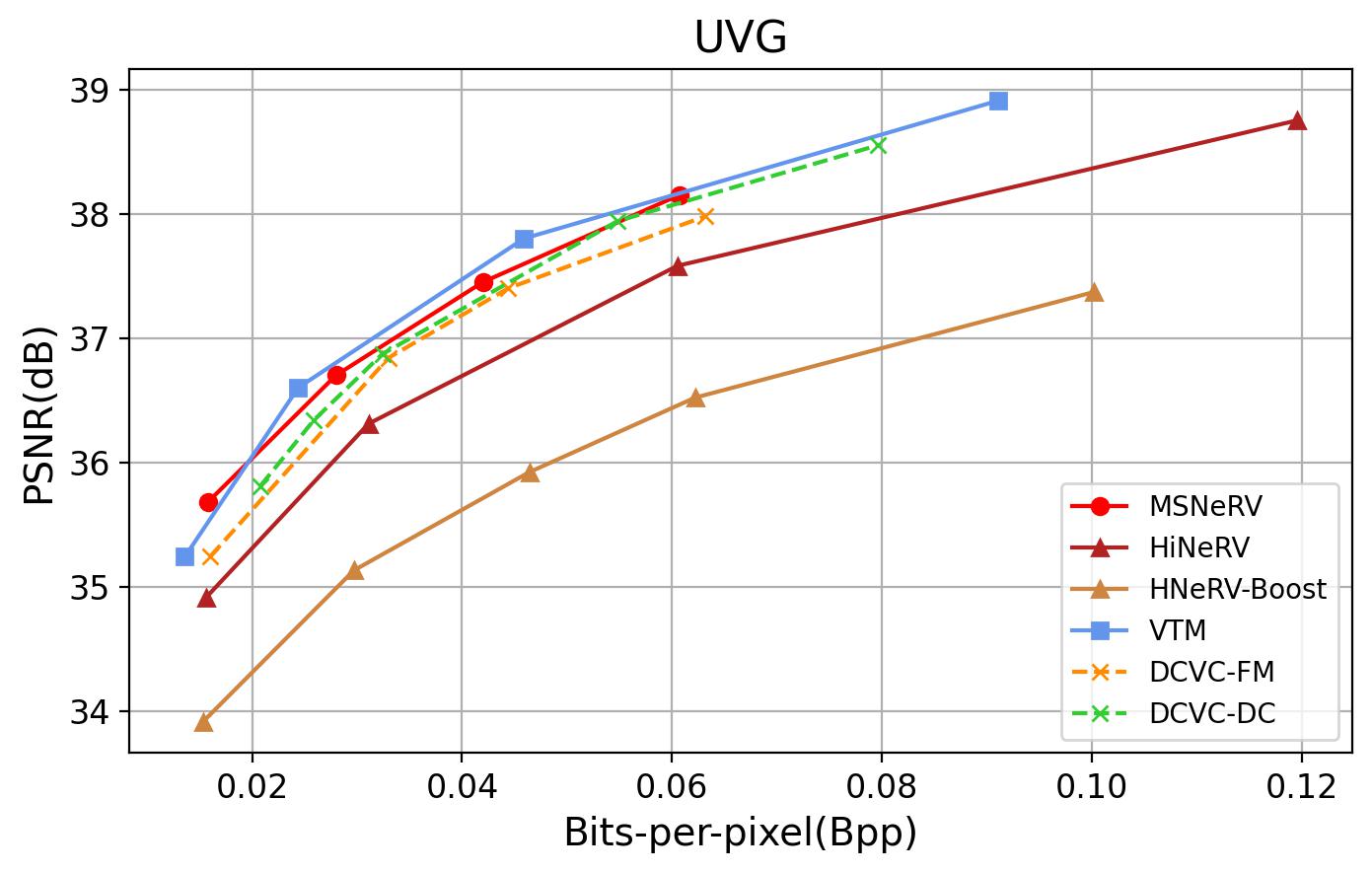}
    \end{subfigure}
    \begin{subfigure}{0.4\textwidth}
        \centering
        \includegraphics[width=\linewidth]{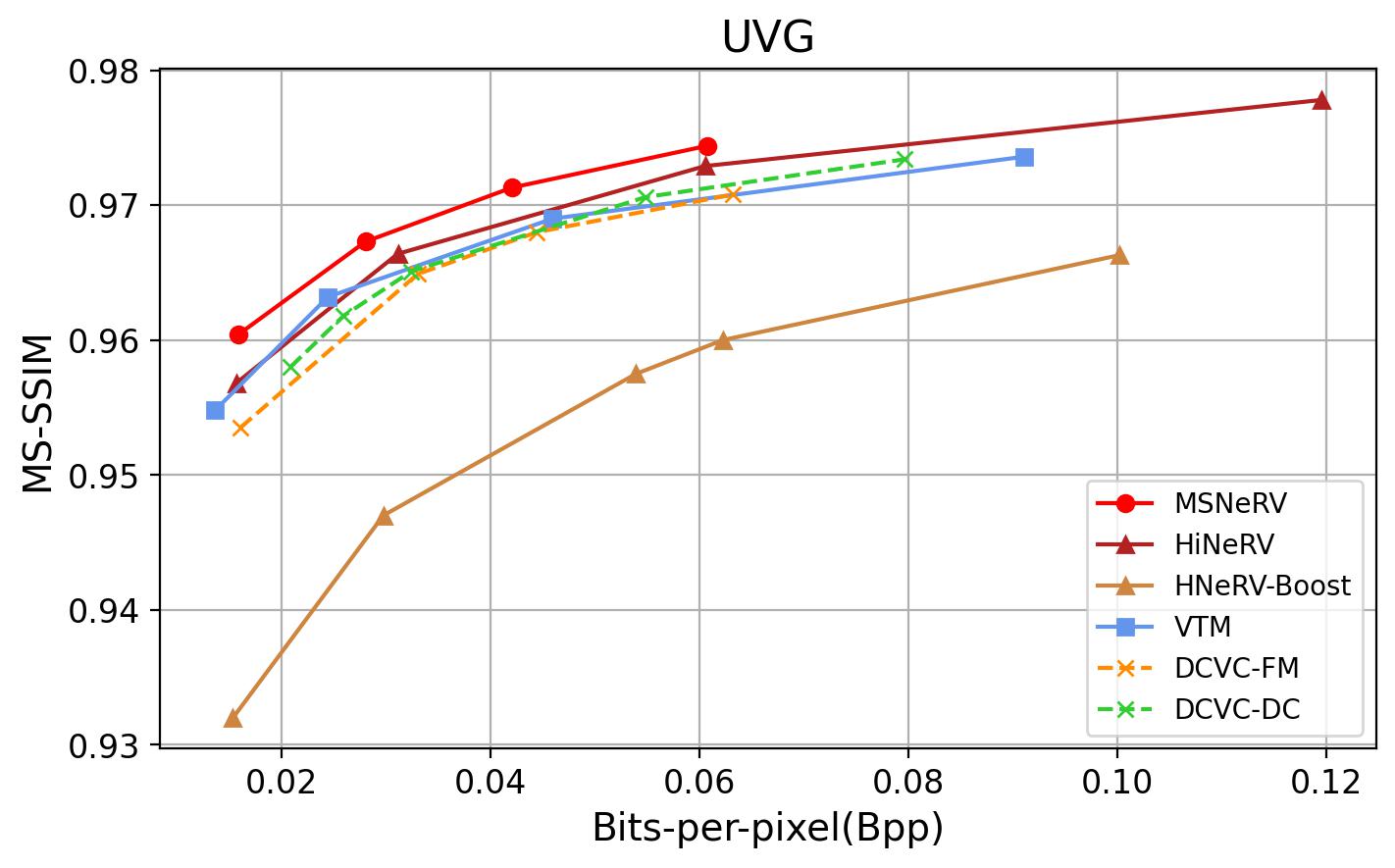}
    \end{subfigure}

    \caption{
Video compression results on the HEVC ClassB \cite{sullivan2012overview} and UVG datasets \cite{mercat2020uvg}.}
    \label{fig:performance}
\end{figure*}

%% file: compare.tex
\begin{figure*}[tp]
    \centering
    \includegraphics[width=\linewidth]{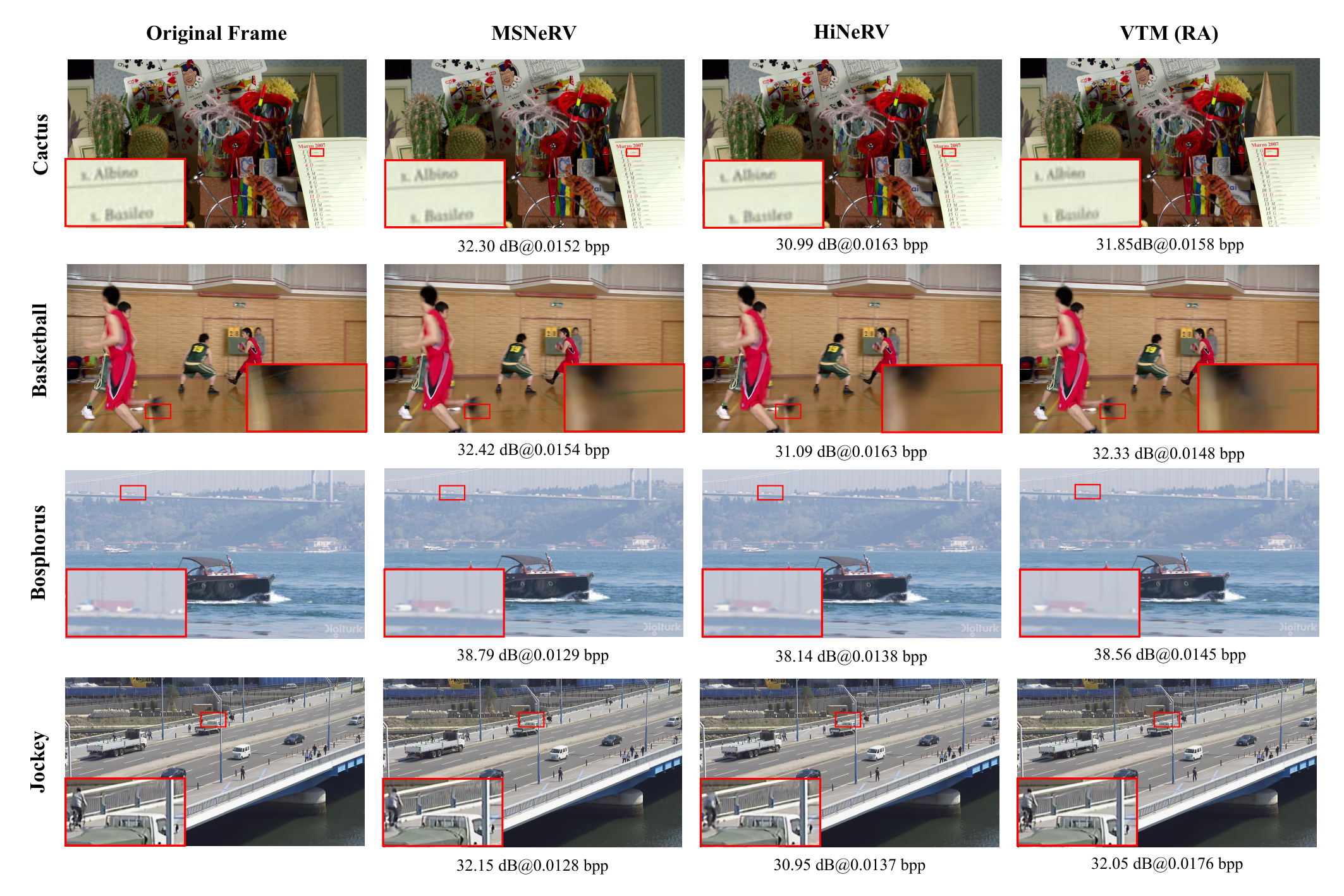}
    \caption{Compression results on different videos. }
    \label{fig:compare}
\end{figure*}

%% file: tables/ablation.tex
\begin{table}[t]
\centering
 \resizebox{\columnwidth}{!}{%
\begin{tabular}{c c c c c}
\specialrule{1.5pt}{0pt}{2pt}
Model & Size (M) & bpp $ \downarrow$ & PSNR(dB) $\uparrow$ & MS-SSIM $\uparrow$\\
\specialrule{1pt}{2pt}{2pt}\\
MSNeRV & 3.28&0.0233&\textbf{33.24}&\textbf{0.9634}\\
V1 & 3.28&0.0236 &33.09 &0.9623 \\
V2 & 3.28&0.0233 &33.08 &0.9631 \\
V3 & 3.28&0.0234 &33.14 &0.9629 \\
V4 & 3.28&0.0231 &33.13 &0.9627 \\
V5 & 3.28&0.0233 &33.01 &0.9617 \\
V6 & 3.25&\textbf{0.0228} &32.98 &0.9613 \\
V7 & 3.27&0.0230& 33.10& 0.9625 \\
\specialrule{1pt}{2pt}{0pt}
\end{tabular}

 }
\caption{
Ablation studies of MSNeRV on the HEVC ClassB \cite{sullivan2012overview} dataset.
}
\vspace{-0.03in}
\label{tab:ablation}
\end{table}

%% file: ent.tex
\begin{figure}[tp]
    \centering
    \includegraphics[width=\linewidth]{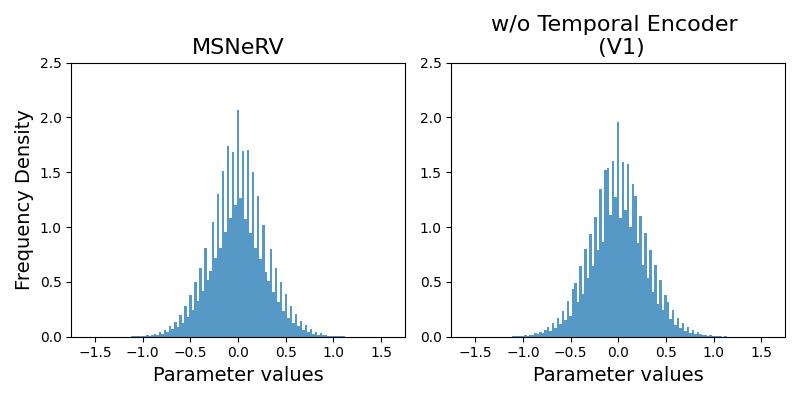}
    \caption{Parameter distributions of MSNeRV (Standard Deviation = 0.290, Estimated Entropy = 6.04) and V1 (Standard Deviation = 0.293, Estimated Entropy = 6.07), offering insights into their differences in parameter allocation.}
    \label{fig:ent}
\end{figure}

%% file: 10_conclusion.tex
\section{Conclusion}
\label{sec:conclusion}
In this paper, we propose a novel neural video representation, MSNeRV, to address the limited fitting ability of INR-based methods for complex videos.
MSNeRV focuses on improving representation capability through efficient fusion of multi-scale information.
We model intricate dynamic relationships through temporal multi-scale fusion encoding, improve detail representation with spatial multi-scale supervision, and fully leverage features via multi-scale feature fusion.
Experimental results demonstrate that MSNeRV significantly outperforms existing INR-based methods in video representation, particularly in dynamic scenes.

Furthermore, we plan to integrate advanced entropy coding and quantization schemes to further enhance video compression performance of our approach.

%% file: main.bbl

%% file: 12_appendix.tex
\section{MSNeRV Architecture}
\label{sec:appendix_section}
At the encoder, the grid size is set to $T\times\frac{H}{24}\times\frac{W}{24}\times C_0$, where $C_0$ denotes the number of channels, which is tied to the model size.
The window size $l$ is set to $5$, with an overlap size of $4$ between adjacent frames.
The Group of Pictures (GoP) is defined by the total number of video frames. Typically, we select $\frac{N}{5}$ frames as a GoP, where $N$ is the total number of frames.
In the decoder, we employ $4$ multi-scale feature blocks with upsampling factors of (3, 2, 2, 2).
Each block includes 3 fusion layers to enhance the feature representation. Additionally, a convolution layer is applied at the final stage to restore the extracted features back to the video domain.
The specific values of $\alpha$ and $\beta$ in the SA loss function are provided in Table \ref{tab:SAloss}.
\input{tables/sup/SAloss}

\section{Implementation Details}
For learning-base methods, we utilize their built-in bitrate control commands to achieve different bpp values.
In HiNeRV and HNeRV, bitrate is adjusted by the number of channels. We observe that deepening the network also improves bitrate. However, this significantly increases GPU memory consumption and training complexity. Therefore, in this paper, we adopt the same channel adjustment strategy to control the bitrate.
In future work, we aim to reduce the network complexity and explore more efficient bitrate adjustment methods.

\input{tables/sup/MRV}
\section{Additional Ablation Results}
\subsection{Video Downsampling}
To generate videos at different resolutions, we explore various downsampling strategies, including average pooling, max pooling, bicubic, and direct subsampling.
We conduct experiments on the ClassB dataset and observe that bicubic and average pooling produce inferior results, as shown in Table \ref{tab:MRA}. 
This is likely due to their inherent smoothing effect during downsampling, which diminishes the prominence of structural information in low-resolution videos.
\subsection{Temporal Window Size}
We explored the performance of different window sizes and multi-layer temporal window stacking strategies. The results in Table \ref{tab:window} show that the best performance is achieved with a single-layer temporal window of size 5.
\input{tables/sup/window}

\subsection{Layer Depths}
We found that adjusting the network’s layer depths can enhance its representational capability.
By reducing high-channel layers and redistributing these parameters to lower-channel layers, the model allocates more resources to detail reconstruction, leading to improved performance. However, this deepens the network, resulting in higher computational complexity, as shown in Table \ref{tab:layer}. We choose the configuration of (3, 3, 3, 3) to achieve a balance between performance and efficiency.

\input{inpaint}
\input{inter2}
\input{tables/sup/layer}
\section{Video Inpainting and Interpolation}
Experiments on video inpainting and interpolation support for the generalization ability of our model.
We evaluate MSNeRV and HiNeRV on the UVG dataset, using video frames with a $120\times120$ central mask as the training set for video inpainting.
For frame interpolation, odd-numbered frames serve as the training set, while even-numbered frames are designated as the test set.
\newpage
We interpolate the grids corresponding to the training set to obtain the required grids for the test set.
These grids are then upsampled to generate the test frames using the trained decoder.
This approach ensures that no information from the test images is introduced, providing a more realistic evaluation consistent with real-world scenarios.
The results are shown in Figure \ref{fig:inpaint}, \ref{fig:inter2}, \ref{fig:inter}.
\input{inter}

%% file: tables/sup/SAloss.tex
\begin{table}[]
\centering
\begin{tabular}{c | c c }
\specialrule{1.5pt}{0pt}{2pt}
Resolution& $\alpha$ & $\beta$ \\
\specialrule{1pt}{2pt}{2pt}
$240\times 135$ (r=1)&0.9&0.1 \\
$480\times 270$ (r=2)&0.8&0.2\\
$960\times 540$ (r=3)&0.7&0.3\\
$1920\times 1080$ (r=4)&0.6&0.3\\

\specialrule{1pt}{2pt}{0pt}
\end{tabular}

\caption{
Values of $\alpha$ and $\beta$ in SA loss.
}
\vspace{-0.05in}
\label{tab:SAloss}
\end{table}

%% file: tables/sup/MRV.tex
\begin{table}[tp]
\centering
\begin{tabular}{c c c c c }
\specialrule{1.5pt}{0pt}{2pt}
 & Bicubic & Direct & Avg-Pooling & Max-Pooling \\
\specialrule{1pt}{2pt}{2pt}
Bas.& 32.46& 32.49& 32.45&\textbf{32.56} \\
BQT.& 32.15& 32.19& 32.17& \textbf{32.23}\\
Cac.& 32.60&32.64& 32.64& \textbf{32.70}\\
Kim.& 35.62&\textbf{35.75}& 35.65& 35.73\\
Par.& 32.63&32.66& 32.63& \textbf{32.71}\\
\specialrule{1pt}{2pt}{2pt}
Avg.& 33.09& 33.15& 33.11&\textbf{33.19}\\
\specialrule{1pt}{2pt}{0pt}
\end{tabular}

\caption{
Results of different downsampling functions.
}
\vspace{-0.05in}
\label{tab:MRA}
\end{table}

%% file: tables/sup/window.tex
\begin{table}[H]
\centering
\begin{tabular}{c| c c }
\specialrule{1.5pt}{0pt}{2pt}
 Window Sizes& Bpp$\downarrow$ & PSNR $\uparrow$\\
\specialrule{1pt}{2pt}{2pt}
3 & \textbf{0.0232}& 33.14\\
5 & 0.0233& \textbf{33.19}\\
7 & 0.0235& 33.18\\
3+3& 0.0233&33.12\\
3+5& 0.0236 & 33.14\\
\specialrule{1pt}{2pt}{0pt}
\end{tabular}

\caption{
Results of different temporal window sizes.
}
\vspace{-0.05in}
\label{tab:window}
\end{table}

%% file: inpaint.tex
\begin{figure*}[h]
    \centering
    \includegraphics[width=\linewidth]{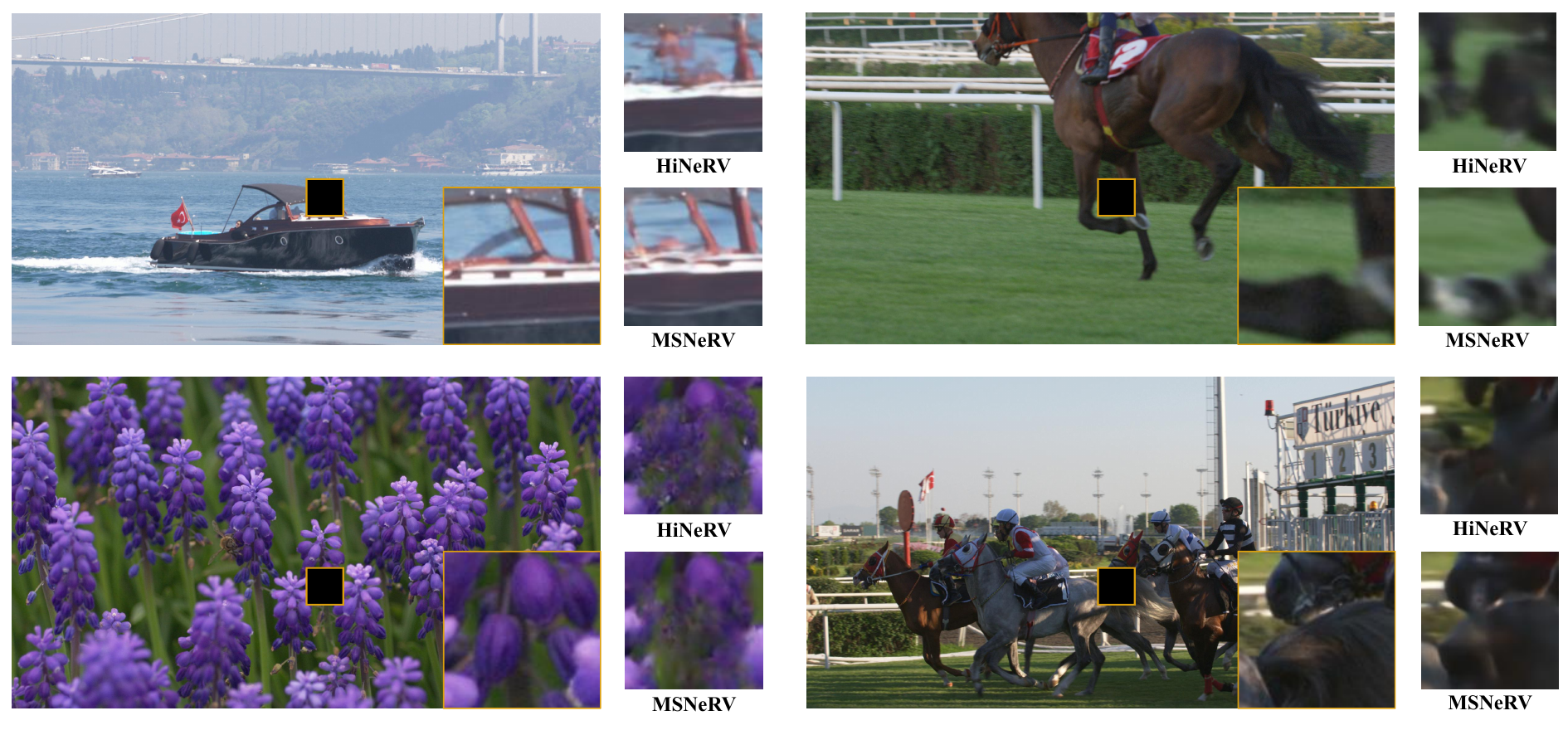}
    \caption{Inpainting results. }
    \label{fig:inpaint}
\end{figure*}

%% file: inter2.tex
\begin{figure*}[h]
    \centering
    \includegraphics[width=\linewidth]{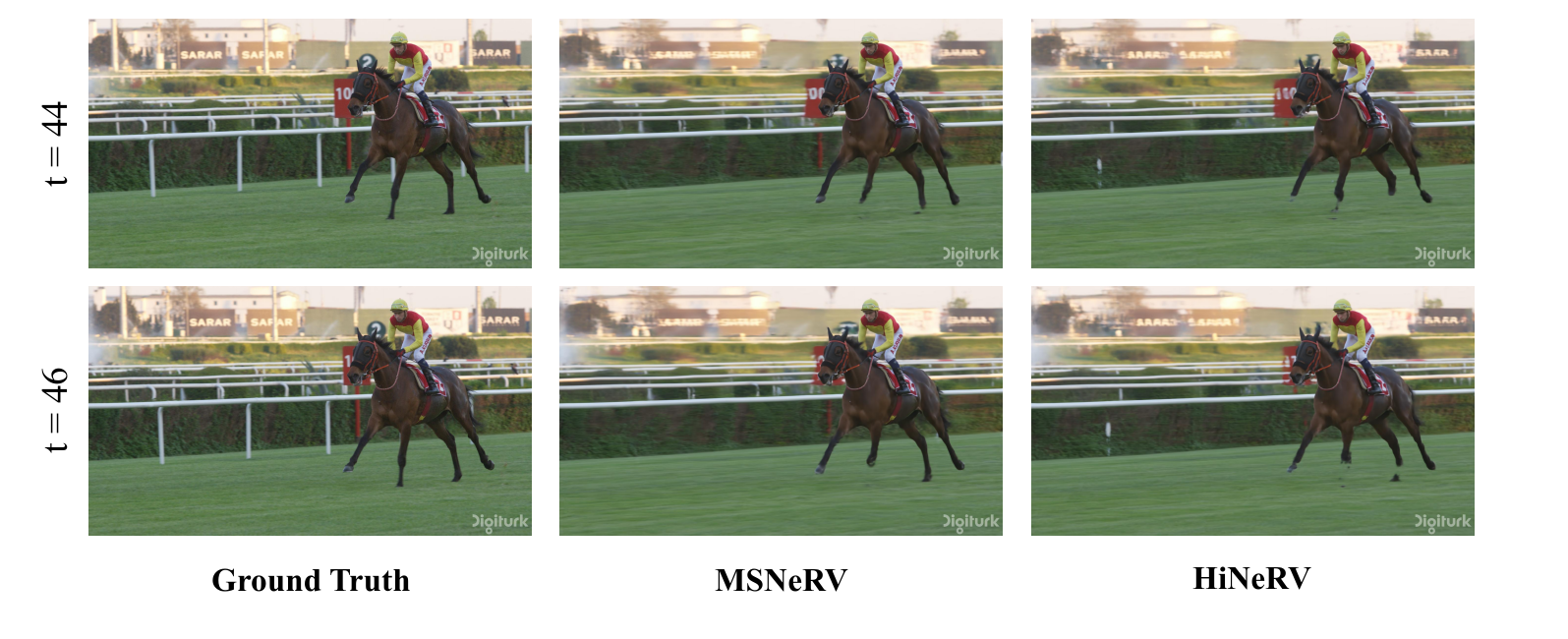}
    \caption{Interpolation results. }
    \label{fig:inter2}
\end{figure*}

%% file: tables/sup/layer.tex
\begin{table}[H]
\centering
\begin{tabular}{c | c c c}
\specialrule{1.5pt}{0pt}{2pt}
Layer Depths & Bpp & PSNR & MACs\\
\specialrule{1pt}{2pt}{2pt}
4, 1, 1, 1 & 0.0173 &30.59 & 108.67 G\\
3, 3, 3, 3 & 0.0163 &31.00 & 181.86 G\\
3, 2, 4, 6 & 0.0158 & 30.96 & 210.70 G\\
2, 6, 6, 7 & 0.0164 & 31.30 & 330.59 G\\
2, 5, 8, 8 & 0.0162 & 31.28 & 357.14 G\\
\specialrule{1pt}{2pt}{0pt}
\end{tabular}

\caption{
Results of different layer depths.
}
\vspace{-0.05in}
\label{tab:layer}
\end{table}

%% file: inter.tex
\begin{figure*}[]
    \centering
    \includegraphics[width=\linewidth]{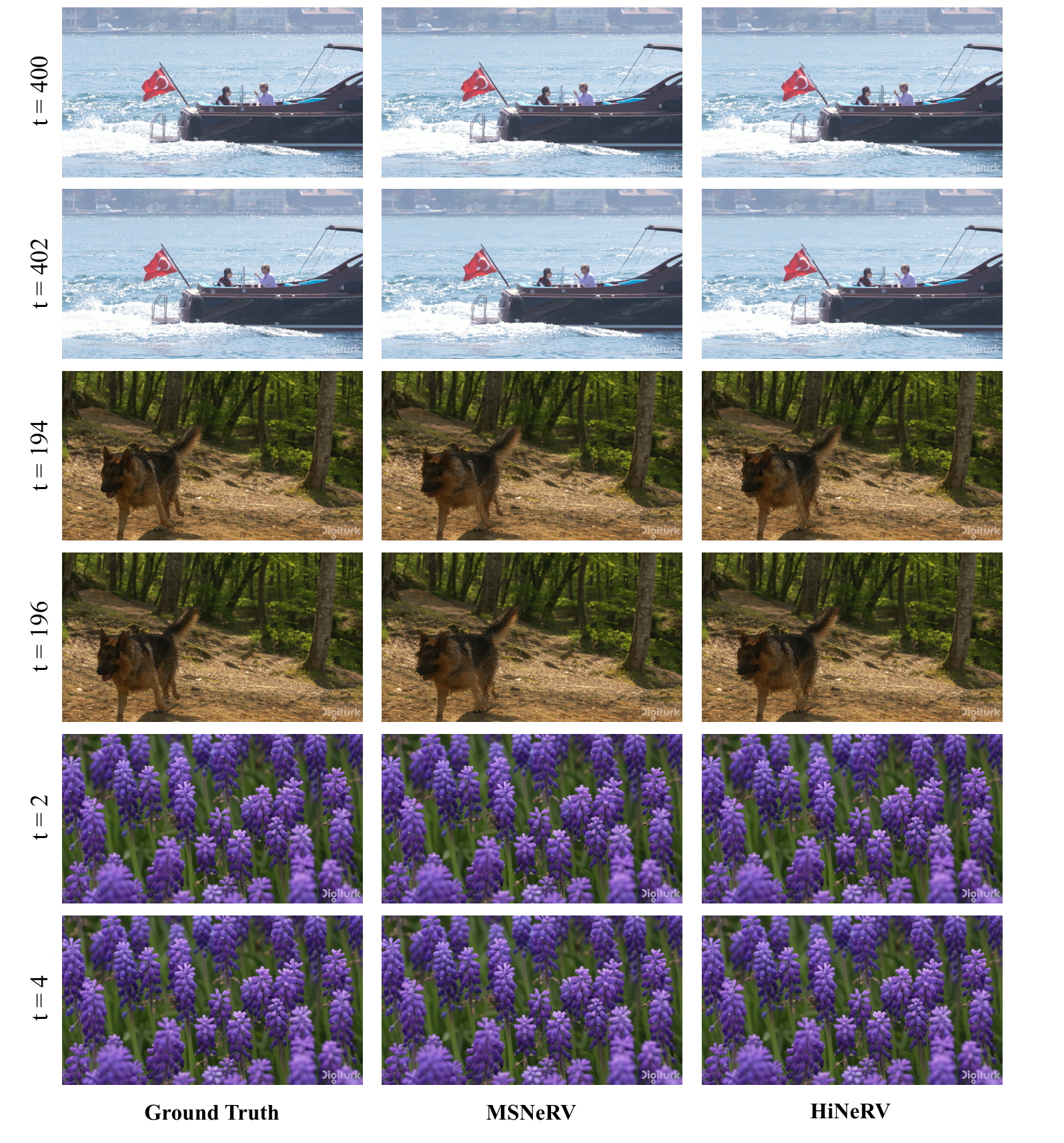}
    \caption{Interpolation results. }
    \label{fig:inter}
\end{figure*}

%% file: _main.bbl
\begin{thebibliography}{10}
\providecommand{\url}[1]{#1}
\csname url@samestyle\endcsname
\providecommand{\newblock}{\relax}
\providecommand{\bibinfo}[2]{#2}
\providecommand{\BIBentrySTDinterwordspacing}{\spaceskip=0pt\relax}
\providecommand{\BIBentryALTinterwordstretchfactor}{4}
\providecommand{\BIBentryALTinterwordspacing}{\spaceskip=\fontdimen2\font plus
\BIBentryALTinterwordstretchfactor\fontdimen3\font minus \fontdimen4\font\relax}
\providecommand{\BIBforeignlanguage}[2]{{%
\expandafter\ifx\csname l@#1\endcsname\relax
\typeout{** WARNING: IEEEtran.bst: No hyphenation pattern has been}%
\typeout{** loaded for the language `#1'. Using the pattern for}%
\typeout{** the default language instead.}%
\else
\language=\csname l@#1\endcsname
\fi
#2}}
\providecommand{\BIBdecl}{\relax}
\BIBdecl

\bibitem{mildenhall2021nerf}
B.~Mildenhall, P.~P. Srinivasan, M.~Tancik, J.~T. Barron, R.~Ramamoorthi, and R.~Ng, ``Nerf: Representing scenes as neural radiance fields for view synthesis,'' \emph{Communications of the ACM}, vol.~65, no.~1, pp. 99--106, 2021.

\bibitem{chen2021nerv}
H.~Chen, B.~He, H.~Wang, Y.~Ren, S.~N. Lim, and A.~Shrivastava, ``Nerv: Neural representations for videos,'' \emph{Advances in Neural Information Processing Systems}, vol.~34, pp. 21\,557--21\,568, 2021.

\bibitem{chen2023hnerv}
H.~Chen, M.~Gwilliam, S.-N. Lim, and A.~Shrivastava, ``Hnerv: A hybrid neural representation for videos,'' in \emph{Proceedings of the IEEE/CVF Conference on Computer Vision and Pattern Recognition}, 2023, pp. 10\,270--10\,279.

\bibitem{kwan2024hinerv}
H.~M. Kwan, G.~Gao, F.~Zhang, A.~Gower, and D.~Bull, ``Hinerv: Video compression with hierarchical encoding-based neural representation,'' \emph{Advances in Neural Information Processing Systems}, vol.~36, 2024.

\bibitem{lu2019dvc}
G.~Lu, W.~Ouyang, D.~Xu, X.~Zhang, C.~Cai, and Z.~Gao, ``Dvc: An end-to-end deep video compression framework,'' in \emph{Proceedings of the IEEE/CVF conference on computer vision and pattern recognition}, 2019, pp. 11\,006--11\,015.

\bibitem{li2021deep}
J.~Li, B.~Li, and Y.~Lu, ``Deep contextual video compression,'' \emph{Advances in Neural Information Processing Systems}, vol.~34, pp. 18\,114--18\,125, 2021.

\bibitem{li2023neural}
------, ``Neural video compression with diverse contexts,'' in \emph{Proceedings of the IEEE/CVF Conference on Computer Vision and Pattern Recognition}, 2023, pp. 22\,616--22\,626.

\bibitem{li2024neural}
------, ``Neural video compression with feature modulation,'' in \emph{Proceedings of the IEEE/CVF Conference on Computer Vision and Pattern Recognition}, 2024, pp. 26\,099--26\,108.

\bibitem{kwan2024nvrc}
H.~M. Kwan, G.~Gao, F.~Zhang, A.~Gower, and D.~Bull, ``Nvrc: Neural video representation compression,'' \emph{arXiv preprint arXiv:2409.07414}, 2024.

\bibitem{bross2021overview}
B.~Bross, Y.-K. Wang, Y.~Ye, S.~Liu, J.~Chen, G.~J. Sullivan, and J.-R. Ohm, ``Overview of the versatile video coding (vvc) standard and its applications,'' \emph{IEEE Transactions on Circuits and Systems for Video Technology}, vol.~31, no.~10, pp. 3736--3764, 2021.

\bibitem{moore1991vtm}
S.~Moore, I.~Curthoys, and S.~McCoy, ``Vtm—an image-processing system for measuring ocular torsion,'' \emph{Computer methods and programs in biomedicine}, vol.~35, no.~3, pp. 219--230, 1991.

\bibitem{wien2015high}
M.~Wien, ``High efficiency video coding,'' \emph{Coding Tools and specification}, vol.~24, p.~1, 2015.

\bibitem{chen2018algorithm}
J.~Chen, Y.~Ye, and S.~Kim, ``Algorithm description for versatile video coding and test model 5,'' \emph{JVET-L1002, Joint Video Exploration Team (JVET)}, 2018.

\bibitem{tang2025canerv}
L.~Tang, J.~Zhu, X.~Zhang, L.~Zhang, S.~Ma, and Q.~Huang, ``Canerv: Content adaptive neural representation for video compression,'' \emph{arXiv preprint arXiv:2502.06181}, 2025.

\bibitem{lee2023ffnerv}
J.~C. Lee, D.~Rho, J.~H. Ko, and E.~Park, ``Ffnerv: Flow-guided frame-wise neural representations for videos,'' in \emph{Proceedings of the 31st ACM International Conference on Multimedia}, 2023, pp. 7859--7870.

\bibitem{sullivan2012overview}
G.~J. Sullivan, J.-R. Ohm, W.-J. Han, and T.~Wiegand, ``Overview of the high efficiency video coding (hevc) standard,'' \emph{IEEE Transactions on circuits and systems for video technology}, vol.~22, no.~12, pp. 1649--1668, 2012.

\bibitem{mercat2020uvg}
A.~Mercat, M.~Viitanen, and J.~Vanne, ``Uvg dataset: 50/120fps 4k sequences for video codec analysis and development,'' in \emph{Proceedings of the 11th ACM multimedia systems conference}, 2020, pp. 297--302.

\bibitem{wiegand2003overview}
T.~Wiegand, G.~J. Sullivan, G.~Bjontegaard, and A.~Luthra, ``Overview of the h. 264/avc video coding standard,'' \emph{IEEE Transactions on circuits and systems for video technology}, vol.~13, no.~7, pp. 560--576, 2003.

\bibitem{laude2016deep}
T.~Laude and J.~Ostermann, ``Deep learning-based intra prediction mode decision for hevc,'' in \emph{2016 Picture Coding Symposium (PCS)}.\hskip 1em plus 0.5em minus 0.4em\relax IEEE, 2016, pp. 1--5.

\bibitem{chen2017deepcoder}
T.~Chen, H.~Liu, Q.~Shen, T.~Yue, X.~Cao, and Z.~Ma, ``Deepcoder: A deep neural network based video compression,'' in \emph{2017 IEEE Visual Communications and Image Processing (VCIP)}.\hskip 1em plus 0.5em minus 0.4em\relax IEEE, 2017, pp. 1--4.

\bibitem{liu2021deep}
B.~Liu, Y.~Chen, S.~Liu, and H.-S. Kim, ``Deep learning in latent space for video prediction and compression,'' in \emph{Proceedings of the IEEE/CVF conference on computer vision and pattern recognition}, 2021, pp. 701--710.

\bibitem{theis2017lossy}
L.~Theis, W.~Shi, A.~Cunningham, and F.~Husz{\'a}r, ``Lossy image compression with compressive autoencoders,'' \emph{arXiv preprint arXiv:1703.00395}, 2017.

\bibitem{zhang2024boosting}
X.~Zhang, R.~Yang, D.~He, X.~Ge, T.~Xu, Y.~Wang, H.~Qin, and J.~Zhang, ``Boosting neural representations for videos with a conditional decoder,'' in \emph{Proceedings of the IEEE/CVF Conference on Computer Vision and Pattern Recognition}, 2024, pp. 2556--2566.

\bibitem{sitzmann2020implicit}
V.~Sitzmann, J.~Martel, A.~Bergman, D.~Lindell, and G.~Wetzstein, ``Implicit neural representations with periodic activation functions,'' \emph{Advances in neural information processing systems}, vol.~33, pp. 7462--7473, 2020.

\bibitem{su2022inras}
K.~Su, M.~Chen, and E.~Shlizerman, ``Inras: Implicit neural representation for audio scenes,'' \emph{Advances in Neural Information Processing Systems}, vol.~35, pp. 8144--8158, 2022.

\bibitem{dupont2021coin}
E.~Dupont, A.~Goli{\'n}ski, M.~Alizadeh, Y.~W. Teh, and A.~Doucet, ``Coin: Compression with implicit neural representations,'' \emph{arXiv preprint arXiv:2103.03123}, 2021.

\bibitem{strumpler2022implicit}
Y.~Str{\"u}mpler, J.~Postels, R.~Yang, L.~V. Gool, and F.~Tombari, ``Implicit neural representations for image compression,'' in \emph{European Conference on Computer Vision}.\hskip 1em plus 0.5em minus 0.4em\relax Springer, 2022, pp. 74--91.

\bibitem{kim2024c3}
H.~Kim, M.~Bauer, L.~Theis, J.~R. Schwarz, and E.~Dupont, ``C3: High-performance and low-complexity neural compression from a single image or video,'' in \emph{Proceedings of the IEEE/CVF Conference on Computer Vision and Pattern Recognition}, 2024, pp. 9347--9358.

\bibitem{pumarola2021d}
A.~Pumarola, E.~Corona, G.~Pons-Moll, and F.~Moreno-Noguer, ``D-nerf: Neural radiance fields for dynamic scenes,'' in \emph{Proceedings of the IEEE/CVF conference on computer vision and pattern recognition}, 2021, pp. 10\,318--10\,327.

\bibitem{song2023nerfplayer}
L.~Song, A.~Chen, Z.~Li, Z.~Chen, L.~Chen, J.~Yuan, Y.~Xu, and A.~Geiger, ``Nerfplayer: A streamable dynamic scene representation with decomposed neural radiance fields,'' \emph{IEEE Transactions on Visualization and Computer Graphics}, vol.~29, no.~5, pp. 2732--2742, 2023.

\bibitem{kuanar2018deep}
S.~Kuanar, C.~Conly, and K.~Rao, ``Deep learning based hevc in-loop filtering for decoder quality enhancement,'' in \emph{2018 Picture Coding Symposium (PCS)}.\hskip 1em plus 0.5em minus 0.4em\relax IEEE, 2018, pp. 164--168.

\bibitem{he2016deep}
K.~He, X.~Zhang, S.~Ren, and J.~Sun, ``Deep residual learning for image recognition,'' in \emph{Proceedings of the IEEE conference on computer vision and pattern recognition}, 2016, pp. 770--778.

\bibitem{lecun1998gradient}
Y.~LeCun, L.~Bottou, Y.~Bengio, and P.~Haffner, ``Gradient-based learning applied to document recognition,'' \emph{Proceedings of the IEEE}, vol.~86, no.~11, pp. 2278--2324, 1998.

\bibitem{jahne2005digital}
B.~J{\"a}hne, \emph{Digital image processing}.\hskip 1em plus 0.5em minus 0.4em\relax Springer Science \& Business Media, 2005.

\bibitem{yan2024ds}
H.~Yan, Z.~Ke, X.~Zhou, T.~Qiu, X.~Shi, and D.~Jiang, ``Ds-nerv: Implicit neural video representation with decomposed static and dynamic codes,'' in \emph{Proceedings of the IEEE/CVF Conference on Computer Vision and Pattern Recognition}, 2024, pp. 23\,019--23\,029.

\bibitem{guo2025metanerv}
J.~Guo, J.~Yao, Z.~Wang, J.~Bu, H.~Wang \emph{et~al.}, ``Metanerv: Meta neural representations for videos with spatial-temporal guidance,'' \emph{arXiv preprint arXiv:2501.02427}, 2025.

\bibitem{zhao2016loss}
H.~Zhao, O.~Gallo, I.~Frosio, and J.~Kautz, ``Loss functions for image restoration with neural networks,'' \emph{IEEE Transactions on computational imaging}, vol.~3, no.~1, pp. 47--57, 2016.

\bibitem{janocha2017loss}
K.~Janocha and W.~M. Czarnecki, ``On loss functions for deep neural networks in classification,'' \emph{arXiv preprint arXiv:1702.05659}, 2017.

\bibitem{shi2016real}
W.~Shi, J.~Caballero, F.~Husz{\'a}r, J.~Totz, A.~P. Aitken, R.~Bishop, D.~Rueckert, and Z.~Wang, ``Real-time single image and video super-resolution using an efficient sub-pixel convolutional neural network,'' in \emph{Proceedings of the IEEE conference on computer vision and pattern recognition}, 2016, pp. 1874--1883.

\bibitem{minnen2020channel}
D.~Minnen and S.~Singh, ``Channel-wise autoregressive entropy models for learned image compression,'' in \emph{2020 IEEE International Conference on Image Processing (ICIP)}.\hskip 1em plus 0.5em minus 0.4em\relax IEEE, 2020, pp. 3339--3343.

\bibitem{zhang2019recursive}
S.~Zhang, Z.~Fan, N.~Ling, and M.~Jiang, ``Recursive residual convolutional neural network-based in-loop filtering for intra frames,'' \emph{IEEE Transactions on Circuits and Systems for Video Technology}, vol.~30, no.~7, pp. 1888--1900, 2019.

\bibitem{szatkowski2022hypersound}
F.~Szatkowski, K.~J. Piczak, P.~Spurek, J.~Tabor, and T.~Trzci{\'n}ski, ``Hypersound: Generating implicit neural representations of audio signals with hypernetworks,'' \emph{arXiv preprint arXiv:2211.01839}, 2022.

\bibitem{xu2018impact}
J.~Xu, B.~Zhou, C.~Zhang, N.~Ke, W.~Jin, and S.~Hao, ``The impact of bitrate and gop pattern on the video quality of h. 265/hevc compression standard,'' in \emph{2018 IEEE International Conference on Signal Processing, Communications and Computing (ICSPCC)}.\hskip 1em plus 0.5em minus 0.4em\relax IEEE, 2018, pp. 1--5.

\bibitem{kingma2014adam}
D.~P. Kingma and J.~Ba, ``Adam: A method for stochastic optimization,'' \emph{arXiv preprint arXiv:1412.6980}, 2014.

\bibitem{jacob2018quantization}
B.~Jacob, S.~Kligys, B.~Chen, M.~Zhu, M.~Tang, A.~Howard, H.~Adam, and D.~Kalenichenko, ``Quantization and training of neural networks for efficient integer-arithmetic-only inference,'' in \emph{Proceedings of the IEEE conference on computer vision and pattern recognition}, 2018, pp. 2704--2713.

\bibitem{bjontegaard2008improvements}
G.~Bjontegaard, ``Improvements of the bd-psnr model,'' \emph{VCEG-AI11}, 2008.

\bibitem{kim2024snerv}
J.~Kim, J.~Lee, and J.-W. Kang, ``Snerv: Spectra-preserving neural representation for video,'' in \emph{European Conference on Computer Vision}.\hskip 1em plus 0.5em minus 0.4em\relax Springer, 2024, pp. 332--348.

\bibitem{mentzer2019practical}
F.~Mentzer, E.~Agustsson, M.~Tschannen, R.~Timofte, and L.~V. Gool, ``Practical full resolution learned lossless image compression,'' in \emph{Proceedings of the IEEE/CVF conference on computer vision and pattern recognition}, 2019, pp. 10\,629--10\,638.

\bibitem{bai2023ps}
Y.~Bai, C.~Dong, C.~Wang, and C.~Yuan, ``Ps-nerv: Patch-wise stylized neural representations for videos,'' in \emph{2023 IEEE International Conference on Image Processing (ICIP)}.\hskip 1em plus 0.5em minus 0.4em\relax IEEE, 2023, pp. 41--45.

\bibitem{browne2022algorithm}
A.~Browne, Y.~Ye, and S.~Kim, ``Algorithm description for versatile video coding and test model 17 (vtm 17),'' in \emph{JVET-Z2002}, 2022.

\bibitem{li2022nerv}
Z.~Li, M.~Wang, H.~Pi, K.~Xu, J.~Mei, and Y.~Liu, ``E-nerv: Expedite neural video representation with disentangled spatial-temporal context,'' in \emph{European Conference on Computer Vision}.\hskip 1em plus 0.5em minus 0.4em\relax Springer, 2022, pp. 267--284.

\bibitem{Tang_2023_ICCV}
L.~Tang, X.~Zhang, G.~Zhang, and X.~Ma, ``Scene matters: Model-based deep video compression,'' in \emph{Proceedings of the IEEE/CVF International Conference on Computer Vision (ICCV)}, October 2023, pp. 12\,481--12\,491.

\bibitem{shen2022nerp}
L.~Shen, J.~Pauly, and L.~Xing, ``Nerp: implicit neural representation learning with prior embedding for sparsely sampled image reconstruction,'' \emph{IEEE Transactions on Neural Networks and Learning Systems}, vol.~35, no.~1, pp. 770--782, 2022.

\bibitem{park2020fast}
S.-h. Park and J.-W. Kang, ``Fast multi-type tree partitioning for versatile video coding using a lightweight neural network,'' \emph{IEEE Transactions on Multimedia}, vol.~23, pp. 4388--4399, 2020.

\bibitem{huang2021block}
Y.-W. Huang, J.~An, H.~Huang, X.~Li, S.-T. Hsiang, K.~Zhang, H.~Gao, J.~Ma, and O.~Chubach, ``Block partitioning structure in the vvc standard,'' \emph{IEEE Transactions on Circuits and Systems for Video Technology}, vol.~31, no.~10, pp. 3818--3833, 2021.

\bibitem{yang2019low}
H.~Yang, L.~Shen, X.~Dong, Q.~Ding, P.~An, and G.~Jiang, ``Low-complexity ctu partition structure decision and fast intra mode decision for versatile video coding,'' \emph{IEEE Transactions on Circuits and Systems for Video Technology}, vol.~30, no.~6, pp. 1668--1682, 2019.

\bibitem{dong2021fast}
X.~Dong, L.~Shen, M.~Yu, and H.~Yang, ``Fast intra mode decision algorithm for versatile video coding,'' \emph{IEEE Transactions on Multimedia}, vol.~24, pp. 400--414, 2021.

\bibitem{zhao2019wide}
L.~Zhao, X.~Zhao, S.~Liu, X.~Li, J.~Lainema, G.~Rath, F.~Urban, and F.~Racap{\'e}, ``Wide angular intra prediction for versatile video coding,'' in \emph{2019 Data Compression Conference (DCC)}.\hskip 1em plus 0.5em minus 0.4em\relax IEEE, 2019, pp. 53--62.

\bibitem{zhu2020deep}
L.~Zhu, Y.~Zhang, S.~Wang, S.~Kwong, X.~Jin, and Y.~Qiao, ``Deep learning-based chroma prediction for intra versatile video coding,'' \emph{IEEE Transactions on Circuits and Systems for Video Technology}, vol.~31, no.~8, pp. 3168--3181, 2020.

\end{thebibliography}
